\documentclass[a4paper, 10pt, conference]{ieeeconf} 

\IEEEoverridecommandlockouts                              

\usepackage{graphicx} 
\usepackage{times} 
\usepackage{amsmath} 
\usepackage{amssymb}  
\usepackage{bm}
\graphicspath{{./Figures/}}
\usepackage[caption=false,font=footnotesize]{subfig}
\usepackage{braket}
\usepackage{floatrow}
\usepackage{tikz}
\usepackage{floatrow}
\usepackage{svg}
\usepackage[hidelinks]{hyperref}

\usepackage{booktabs,ragged2e}
\usepackage[flushleft]{threeparttable}

\usepackage{xcolor}

\definecolor{orange}{rgb}{1.0, 0.22, 0.0}

\title{\LARGE \bf
Development of a Novel Impedance-Controlled Quasi-Direct-Drive Robotic Hand}

\author{Jay Best, Amin Fakhari
\thanks{The authors are with the Department of Mechanical Engineering, Stony Brook University, Stony Brook, NY 11794, USA, {\tt\small jaybestiv@gmail.com, amin.fakhari@stonybrook.edu}.}
}

\begin{document}

\maketitle
\thispagestyle{empty}
\pagestyle{empty}

\begin{abstract}
Most robotic hands and grippers rely on actuators with large gearboxes and force sensors for controlling gripping force. However, this might not be ideal for tasks that require the robot to interact with an unstructured and unknown environment. In this paper, we introduce a novel quasi-direct-drive two-fingered robotic hand with variable impedance control in the joint space and Cartesian space. The hand has a total of four degrees of freedom, backdrivable differential gear trains, and four brushless direct current (BLDC) motors. Motor torque is controlled through Field-Oriented Control (FOC) with current sensing. Variable impedance control enables the robotic hand to execute dexterous manipulation tasks safely during environment-robot and human-robot interactions. The quasi-direct-drive actuators eliminate the need for complex tactile/force sensors or precise motion planning when handling environmental contact. A majority-3D-printed assembly makes this a low-cost research platform built with affordable, readily available off-the-shelf components. Experimental validation demonstrates the robotic hand's capability for stable force-closure and form-closure grasps in the presence of disturbances, reliable in-hand manipulation, and safe dynamic manipulations despite contact with the environment.


\end{abstract}

\section{Introduction}
\begin{figure}[!hbtp]
\centering
\includegraphics[scale=0.15]{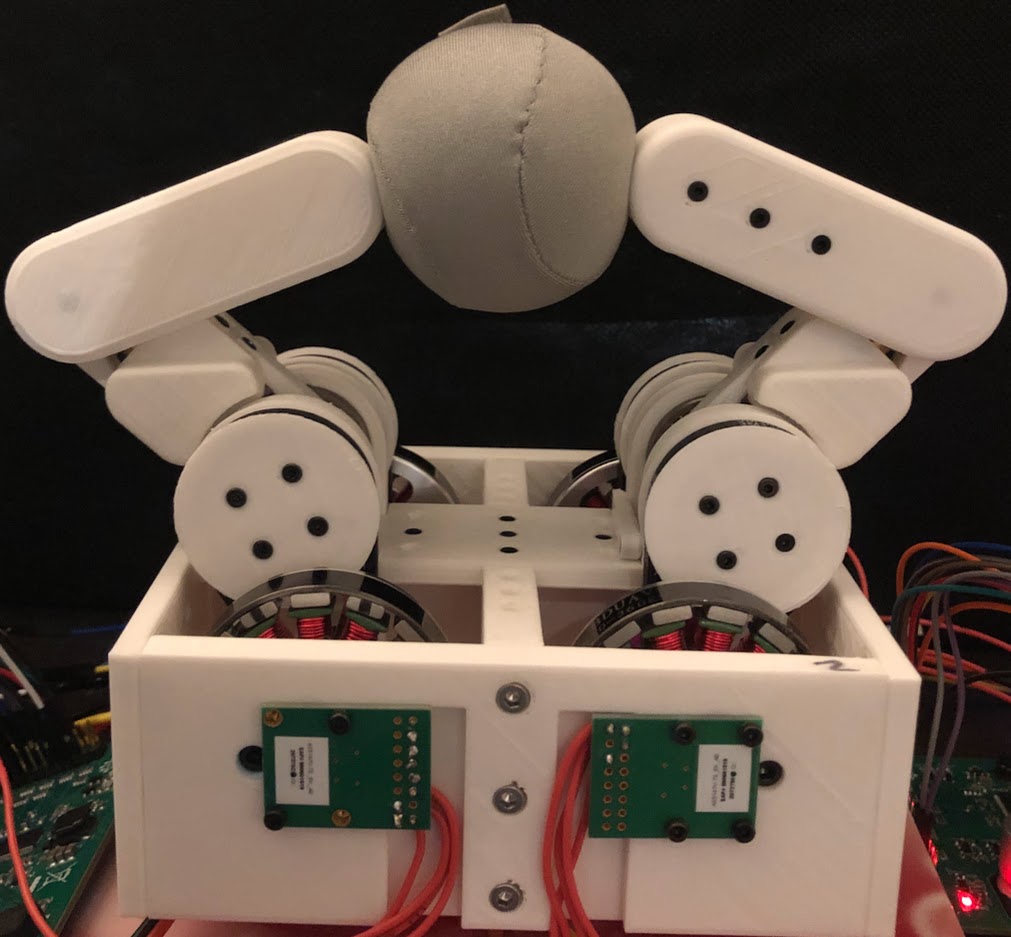}
\caption{The impedance-controlled quasi-direct-drive robotic hand holding a ball.}
\label{fig:intro_pic}
\end{figure}

Robotic grippers and hands have come a long way, however, there are still challenges when they have to significantly interact with an environment, e.g., picking a small object like a coin from an edge of a table, or rapid/dynamic grasping of a small object off an unstructured environment where the impact between the gripper/hand and the environment is inevitable. In these contact-rich grasping and manipulation scenarios, we have to prioritize the adaptivity and stiffness variation capability of the gripper/hand over large gripping forces. Therefore, the gripper will be able to comply with the environment and avoid damaging itself, the object, and possibly the environment.
Different strategies have been employed to handle and control contact of the gripper/hand with the environment during manipulation to ensure a safe and stable grasp and manipulation. One of the strategies that might be useful to explore further is impedance control combined with Direct-Drive (DD) or Quasi-Direct-Drive (QDD) actuators (Fig.~\ref{fig:intro_pic}).

Impedance control differs from more traditional manipulation approaches in that the relationship between the force and position of the end-effector can be dynamically controlled \cite{Hogan1984}. This is especially important for grasping and dexterous manipulation tasks where contacts with the environment must be effectively dealt with. For example, if we were to apply force to the fingers of a robotic hand, we would want to be able to control how the fingers respond to this force. Impedance control models this system as a simple linear mass-spring-damper system. In this way, instead of rigidly moving the fingers through a set of prescribed positions, we can treat the fingers as a ``virtual spring" with a desired starting position, stiffness, and damping behavior. Displacing the finger from its desired position will create a force in accordance with the behavior of the mass-spring-damper system. This contact force will be used to perform various manipulation tasks.

\begin{figure}[!htbp]
    \centering
    \sidesubfloat[]{\includegraphics[scale=0.4]{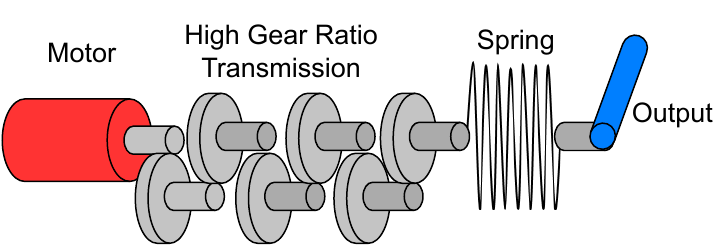}} \quad
    \sidesubfloat[]{\includegraphics[scale=0.4]{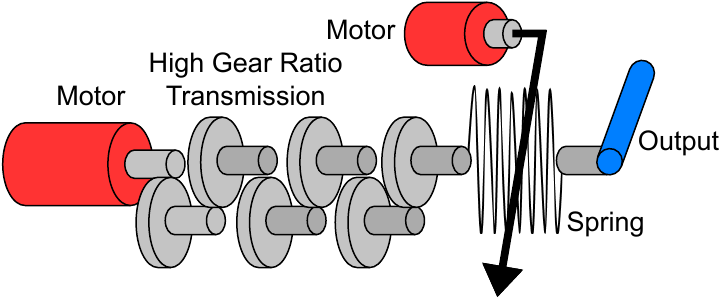}} \\
    \sidesubfloat[]{\includegraphics[scale=0.4]{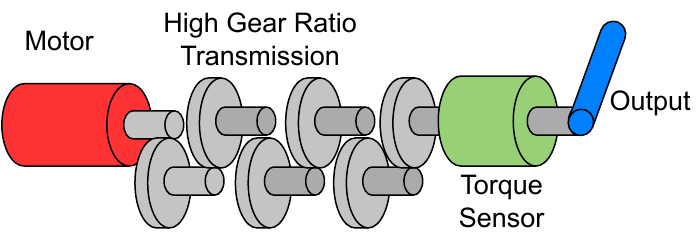}} \quad 
    \sidesubfloat[]{\includegraphics[scale=0.4]{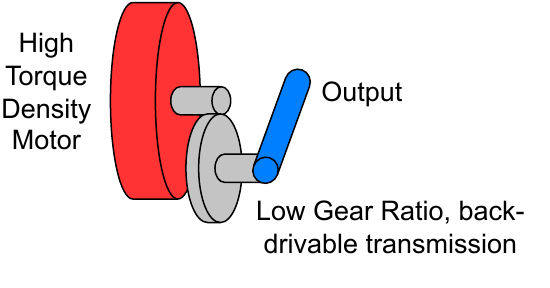}} \\
    \caption{
    Classification of electric drive trains, (a) Series elastic actuators, (b) Variable stiffness actuator, (c) Geared motors with force/torque sensors, (d) Proprioceptive actuators (direct-drive and quasi-direct-drive).}
\label{fig:classification}
\end{figure}

Some of the most common electric drive trains used in robotics are summarized in Fig.~\ref{fig:classification}. Series elastic actuators (Fig.~\ref{fig:classification}-a) work by using a motor with a large gear reduction and a spring placed in between the motor and output shaft. By measuring the deformation of the spring with a sensor, the output torque can be calculated and controlled. Variable stiffness actuators (Fig.~\ref{fig:classification}-b) are a variation of the previous concept with the addition of an additional motor to control the stiffness of the spring. Both of these methods can accurately control torque but at the expense of additional mechanical complexity. Maximum torque is also limited by the stiffness of the spring being used. Geared motors with force/torque sensors (Fig.~\ref{fig:classification}-c) are commonly used in robotic applications.
The advantage of this is that by using a large gear reduction, the maximum output force of the end-effector can be very high, along with the overall actuator torque density. However, force control bandwidth is limited by the torque sensor bandwidth. A large gear reduction also introduces additional complexity and friction in the mechanical design and control of the actuator.
\textit{Proprioceptive} actuators (Fig.~\ref{fig:classification}-d) use a higher torque density motor with a smaller gear reduction, and they include both direct-drive and quasi-direct-drive actuators. By reducing friction in the gear train, the mechanism is highly backdrivable, and motor phase currents can be used to control the output torque. Backdriveability refers to how easily the output of a particular mechanism can be moved around when the motors are powered off. Due to their high bandwidth force control and impact resistance, this category of actuator is used heavily in walking robots such as the MIT Cheetah which experience highly dynamic interaction between the robot legs and the ground \cite{7827048}. For this reason, we believe that this type of actuator would be well suited for handling the contact forces between the fingers in a robotic hand and an object or an unstructured environment. Compared with other methods, these actuators suffer from reduced torque density, however, by using brushless direct current (BLDC) motors with more sophisticated methods of control such as Field-Oriented Control (FOC), we can control torque accurately and more efficiently while maintaining a high enough force output for most basic tasks. The comparative lack of friction in direct-drive and quasi-direct-drive robots makes them a better candidate for applications requiring variable force and compliance control while not requiring external force sensors or springs \cite{10.5555/27675}. However, for quasi-direct-drive actuators to control torque effectively, the gear reduction should be in general less than 10:1 with negligible friction \cite{manipulation}.

\subsection{Related Work}
One of the common forms of variable compliance control in robotic hands and grippers is through force and torque sensors. This includes tactile force sensors at the griper/fingertips and torque sensors in the actuator gear train. Salisbury was one of the first to create a multi-fingered robotic hand with stiffness control using force sensors \cite{salisbury, doi:10.1177/027836498200100102}. The hand had three fingers with 3 DOF each driven by cables enabling the hand to grasp objects in its fingertips and complete some basic in-hand manipulation tasks. Stiffness control as used here is a subset of impedance control which neglects controlling the damping behavior of the end-effector. Another example is the DLR/HIT Hand II \cite{DLR_Hit_II} which used strain gauge-based torque sensors integrated with harmonic gears to measure joint torques used in the control scheme. Joint and Cartesian level impedance controllers are implemented to control finger compliance, while an extended Kalman filter is used for friction estimation in the finger gears. The Ishikawa hand \cite{IshikawaHand} also uses strain gauge-based torque sensors with harmonic gearboxes for force control, and it is able to achieve highly dynamic motion for tasks such as throwing, catching, and dynamic regrasping \cite{Ishikawa_Regrasping}. The two prior examples achieve high performance, however, the introduction of harmonic gearboxes with strain gauges significantly increases cost and mechanical complexity which might not be desirable in certain applications.

Park \textit{et al.} \cite{Park_VariableGraspingStiffness} developed an anthropomorphic robotic hand using Series Elastic Actuator (SEA) modules and fingertip force sensors for tactile feedback. The surface hardness of the object being grasped was estimated by the controller, and finger stiffness was adjusted accordingly through modulating impedance parameters. By this method, a variety of both hard and fragile objects were grasped successfully without damage. Hu \textit{et al.} \cite{hu_liu_xie_yao_liu_2022} created a design for dexterous finger with antagonistic variable stiffness actuators which varied the stiffness of the finger by modulating spring tension in the actuators. This pre-loaded tension on the springs in the finger actuators could be changed to obtain a desired stiffness behavior. The finger achieves 2 DOF by using a differential drive gear system whereby one finger joint is driven by the sum of two actuator torques, and the other finger joint is driven by the difference in actuator torques. This finger mechanism helped serve as an inspiration for our design.

There are some direct-drive and quasi-direct-drive hands and grippers developed in recent years. DDHand \cite{Bhatia2019DirectDH,9981569} uses direct-drive linkages to achieve high bandwidth force control without force sensors. The linkages are easily back-driven, enabling the actuators to effectively act as sensors that sense the environment while picking up an object at an unknown height. They demonstrated the effectiveness of their direct-drive mechanism in a manipulation task called ``smack and snatch" where an object is rapidly grasped off of an unstructured environment. Direct-drive hands lack the high gripping force compared to traditional hands and grippers. However, quasi-direct-drive hands can use a gear ratio of less than 10:1 to retain the benefits of direct-drive with increased torque and force output. Lin \textit{et al.} \cite{QuasiDirectDrive_Lin} developed a novel QDD linkage-based robotic hand with a 4:1 gear reduction. With a QDD actuator, they were able to replicate the "smack and snatch" task first performed by the DDHand. They also integrated a hybrid-drive system with a secondary linear actuator which could be used to provide additional force when the QDD actuators were not sufficient. Another example of QDD manipulation is the TriFinger \cite{TriFinger}, a manipulation platform with three fingers designed from modified QDD actuators from the robot quadruped, Solo. The fingers are driven by BLDC motors with a 9:1 gear reduction allowing the mechanism to be backdrivable and torque controlled with current feedback. The transparent belt reduction allows the fingers to be impact-resistant and run for long periods without damaging hardware. One last example is the manipulation platform Blue \cite{8843134,QuasiDirectDrive_Gealy}, a quasi-direct-drive 7 DOF robot arm driven by low gear reduction BLDC motors complete with a parallel jaw gripper. While parallel jaw grippers can perform well, they may have limited functionality/dexterity when it comes to certain tasks, such as in-hand manipulation when compared to robotic hands with multi-degree-of-freedom fingers.

Our design presents design alternatives and potential improvements to existing robotic hands and grippers within the proprioceptive actuator category. Compared with the other direct-drive and quasi-direct-drive hands \cite{Bhatia2019DirectDH,9981569,QuasiDirectDrive_Lin,TriFinger}, our finger links are more anthropomorphic in nature. This is beneficial for performing a wide variety of grasps such as both force closure and form closure, which are not possible with hands resembling parallel jaw grippers. Our design is capable of performing in-hand manipulation tasks by manipulating objects in the fingertips. Implementing impedance control in the joint space or Cartesian space allows the hand to safely handle contacts with both objects and the environment, while also maintaining stable grasps in response to external wrenches on the object being grasped. For these reasons, we believe this design may serve as an inspiration for future quasi-direct-drive general-purpose robotic hands for use in industry, humanoid robots, prostheses, and potentially other applications.

The contributions of our robotic hand design are as follows:
\begin{itemize}
  \item Integration of a quasi-direct-drive actuation scheme with a differential drive gear train to control movements of 2 DOF fingers.
  \item Implementation of impedance control in joint space and Cartesian spaces using field-oriented control to control torque for force closure grasp, form closure grasp, in-hand manipulation, and safe contact with an unstructured environment.
\end{itemize}


\section{Theoretical Background}

\subsection{Impedance Control Methodology}
Impedance control models the actuator as a spring mass damper system with the adjustable parameters being desired position, stiffness coefficient, and damping coefficient. For a multi-DOF finger/arm,
impedance control can be realized in either the joint space or  Cartesian space.

\begin{figure}[!htbp]
\centering
\subfloat[]{\includegraphics[scale=0.041]{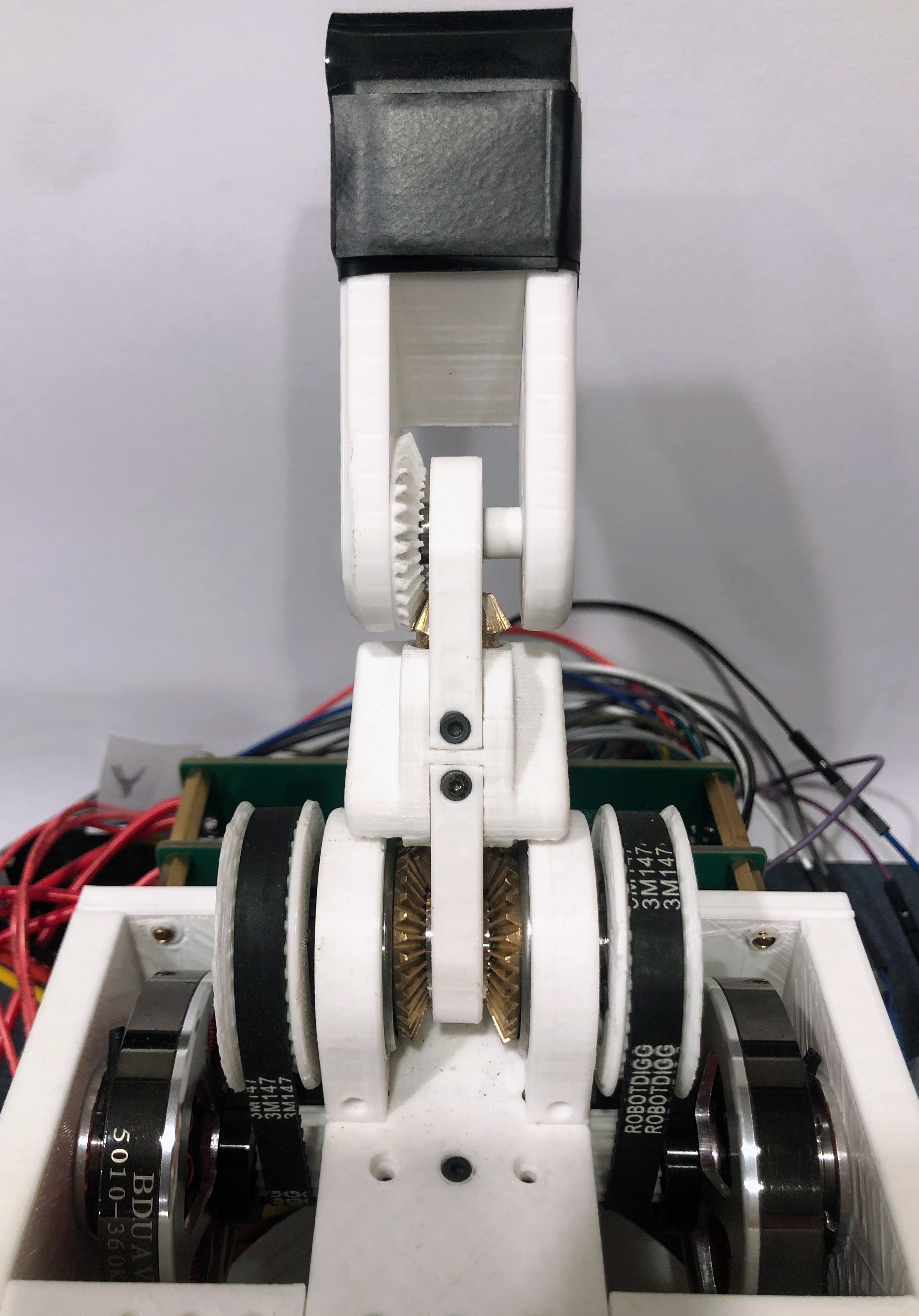}} \qquad
\subfloat[]{\includegraphics[scale=0.45]{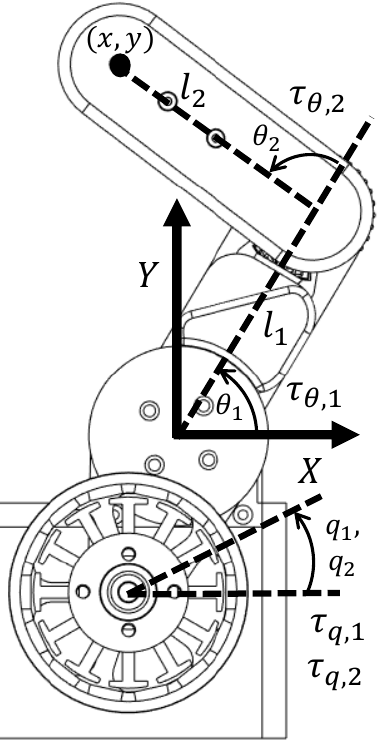}}
\caption{Model of the designed 2-DOF finger with a differential drive gear train, (a) 3D printed model (see Fig.~\ref{fig:HandDrawings} for more details), (b) Kinematic model (including 2 BLDC motors). 
}
\label{fig:Kinematic Model2}
\end{figure}

The 3D printed model and kinematic model of our designed planar 2-DOF finger mechanism are shown in Fig.~\ref{fig:Kinematic Model2}. Since our mechanism uses a differential gear drive to drive the respective joints of each 2-DOF finger, joint angles $\boldsymbol{\theta} =(\theta_1,\theta_2) \in \mathbb{R}^2$ are calculated with motor encoder values $\boldsymbol{q}=(q_1,q_2) \in \mathbb{R}^2$ using \cite{hu_liu_xie_yao_liu_2022} 
\begin{equation}
\boldsymbol{\theta}
=
\begin{bmatrix}
1/(2n_1) & -1/(2n_1) \\
1/(2n_1n_2) & 1/(2n_1n_2)
\end{bmatrix}
\boldsymbol{q},
\label{theta}
\end{equation}
where $n_1$ is the motor belt reduction, which in our case $n_1 = 2.57$, and $n_2$ is the bevel gear reduction in the second joint, which in our case $n_2 = 1$ (see Fig.~\ref{fig:HandDrawings} for more details).

Due to the low inertia and mass of the 3D-printed fingers,
inertia can be neglected. At small velocities, we can calculate joint torques $\boldsymbol{\tau}_\theta =(\tau_{\theta,1},\tau_{\theta,2}) \in \mathbb{R}^2$ for an impedance-controlled finger in Cartesian space using 
\begin{equation}
\boldsymbol{\tau}_\theta = \boldsymbol{J}^\mathrm{T}(\boldsymbol{\theta})\left(\boldsymbol{K}_C(\boldsymbol{x}_d - \boldsymbol{x}) - \boldsymbol{B}_C\dot{\boldsymbol{x}}\right),
\label{impedance4}
\end{equation}
where $\boldsymbol{x}=(x,y) \in \mathbb{R}^2$ is the position of the fingertip that can be computed using the forward kinematics $\boldsymbol{x}=\mathrm{FK}(\boldsymbol{\theta})$, $\boldsymbol{x}_d =(x_d,y_d) \in \mathbb{R}^2$ is the desired position of the fingertip, $\boldsymbol{J}(\boldsymbol{\theta}) \in {\mathbb{R}^{2 \times 2}}$ is the Jacobian matrix where $\dot{\boldsymbol{x}}=\boldsymbol{J}(\boldsymbol{\theta}) \dot{\boldsymbol{\theta}}$, and $\boldsymbol{B}_C\in{\mathbb{R}^{2\times 2}}$ and $\boldsymbol{K}_C\in{\mathbb{R}^{2\times 2}}$ are the damping and stiffness diagonal matrices in Cartesian space, respectively, defined by the user according to a particular task, as
\begin{equation}
\boldsymbol{K}_C=\left[\begin{array}{cc}
k_X & 0 \\
0 & k_Y
\end{array}\right], \qquad \boldsymbol{B}_C=\left[\begin{array}{cc}
b_X & 0 \\
0 & b_Y
\end{array}\right],
\label{eq:K_C_B_C}
\end{equation}
where $k_X$, $k_Y$, $b_X$, and $b_Y$ represent the desired stiffness and damping behavior in the $X$ and $Y$ directions.






Similarly, the joint-space impedance controller can be written as
\begin{equation}
\boldsymbol{\tau}_\theta=\boldsymbol{K}_\theta\left(\boldsymbol{\theta}_d-\boldsymbol{\theta}\right)-\boldsymbol{B}_\theta \dot{\boldsymbol{\theta}},
\label{eq:joint-space impedance controller}
\end{equation}
where $\boldsymbol{B}_\theta \in {\mathbb{R}^{2 \times 2}}$ and $\boldsymbol{K}_\theta \in {\mathbb{R}^{2 \times 2}}$ are the desired damping and stiffness diagonal matrices in joint space, respectively.

Since the mechanism uses a differential gear drive, the finger joint torques $\boldsymbol{\tau}_\theta$, computed by \eqref{impedance4} or \eqref{eq:joint-space impedance controller}, must be converted to the required actuator torques $\boldsymbol{\tau}_q =(\tau_{q,1},\tau_{q,2}) \in \mathbb{R}^2$ by
\begin{equation}
\boldsymbol{\tau}_q=\begin{bmatrix}
1/(2n_1) & 1/(2n_1n_2)\\ 
-1/(2n_1) & 1/(2n_1n_2)
\end{bmatrix}
\boldsymbol{\tau}_\theta.
\label{ta}
\end{equation}


\subsection{Block Diagram of Impedance Controller Implementation}
\begin{figure}[!htbp]
\centering
\sidesubfloat[]{\includegraphics[scale=0.6]{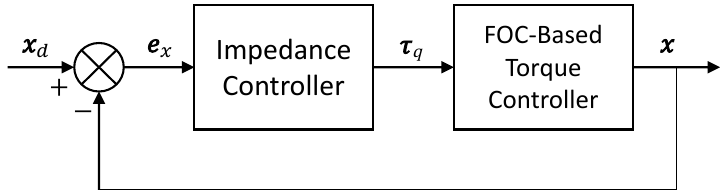}}\\ \vspace{3mm}
\sidesubfloat[]{\includegraphics[scale=0.24]{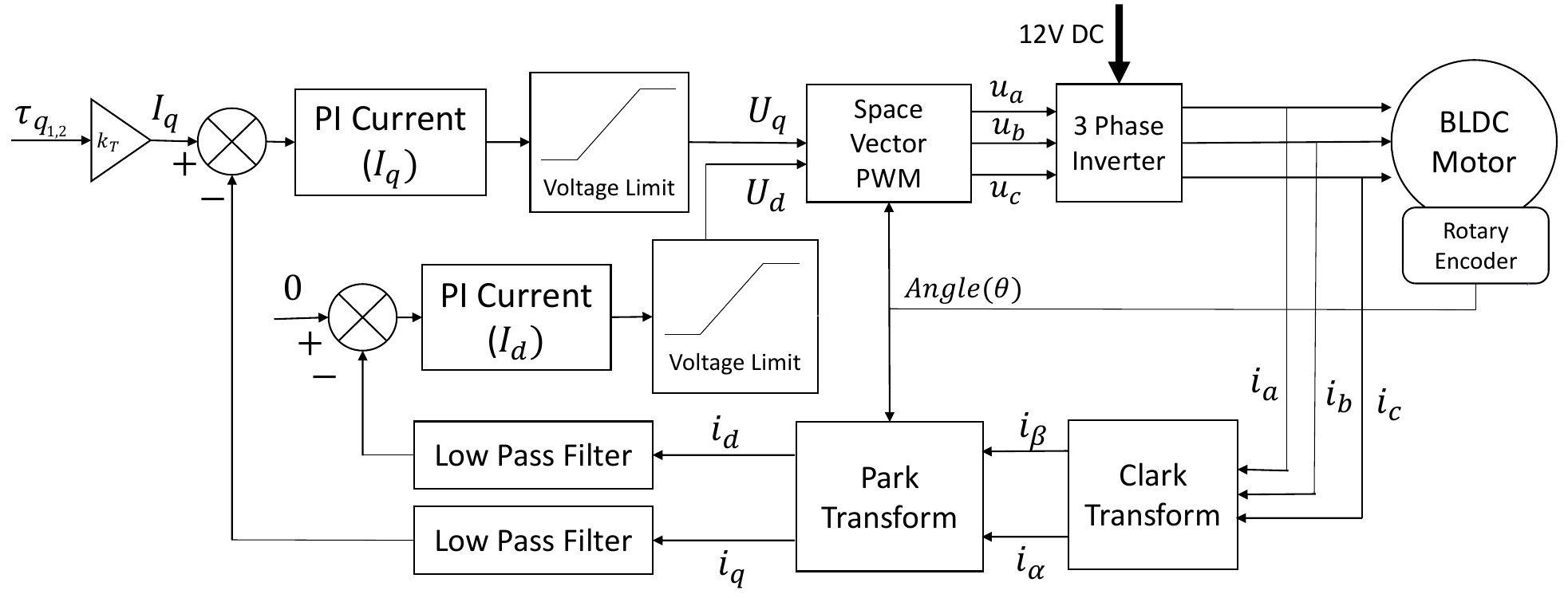}}
\caption{(a) Block diagram of the Cartesian-space impedance controller implementation for each finger, (b) Expanded detailed view of FOC-based torque controller.}
\label{fig:ControllerBlockDiagram}
\end{figure}

The control system for each finger of the robotic hand consists of an outer impedance controller for adjusting the stiffness and damping coefficients, and an inner FOC-based torque controller for each motor to control the torque provided by the impedance controller (Fig.~\ref{fig:ControllerBlockDiagram}). The FOC torque control is implemented by leveraging the SimpleFOC open-source Arduino library \cite{simplefoc2022}. This method enables accurate and efficient torque control at high bandwidth which is essential for dynamic manipulation tasks \cite{act8040071}. This inner torque controller runs 5 times faster than the outer loop impedance controller.


\section{Design Detail}

\subsection{Mechanical Design}
Our approach presents a variety of design challenges. In order to minimize inertia of the finger, all motors should be mounted within the base of the robotic hand, and not within the fingers. Therefore, there must be some way to transfer motion from the motors to the first and second joint of the finger. One way of doing this is to use a tendon/cable mechanism that pulls on the finger joints in order to drive them. However, a disadvantage of this approach is that it is difficult to reduce friction on the cable as it passes through the finger. Since the only torque/force feedback used by the motor is current measurement, friction should be reduced as much as possible, so that the motor torque has a linear relationship with the force output of the finger. Additionally, the gear ratio of the motor should be minimized to preserve the relationship between motor torque and current. Gearing also needs to be backdrivable and compliant to reduce the inherent stiffness of the overall mechanism. Therefore, we chose to use a simple 2.57:1 belt reduction connected to a differential gearbox at the base of the first finger joint. The belt reduction is highly backdrivable, and the bevel gears used in the differential gearbox have relatively low friction.

One problem with a quasi-direct drive, or proprioceptive actuation scheme is that the overall torque density of the actuators tends to be lower when compared to using motors with larger reduction gearboxes. Using a differential drive gearbox as we have implemented helps to compensate for this. The torque at each joint is shared between the two motors for each finger. This allows for higher possible torque at each joint, minimizing the disadvantage that comes with using a smaller gear ratio.

For position feedback, the motor shaft each contains a polarized magnet. The magnetic field is measured by magnetic encoders which send the angular position of each motor to a microcontroller to be used in the impedance control loop, and FOC control loop.

\begin{figure}[htp]
\centering
\sidesubfloat[]{\includegraphics[width= 0.45\textwidth]{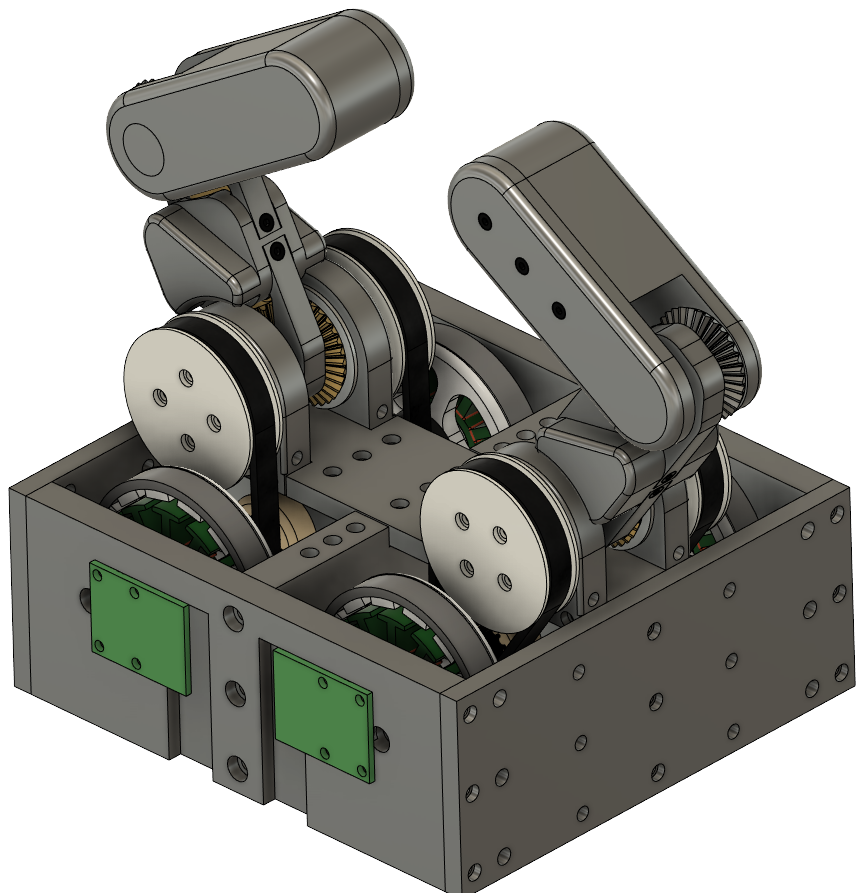}} \,
\sidesubfloat[]{\includegraphics[width=0.4\textwidth]{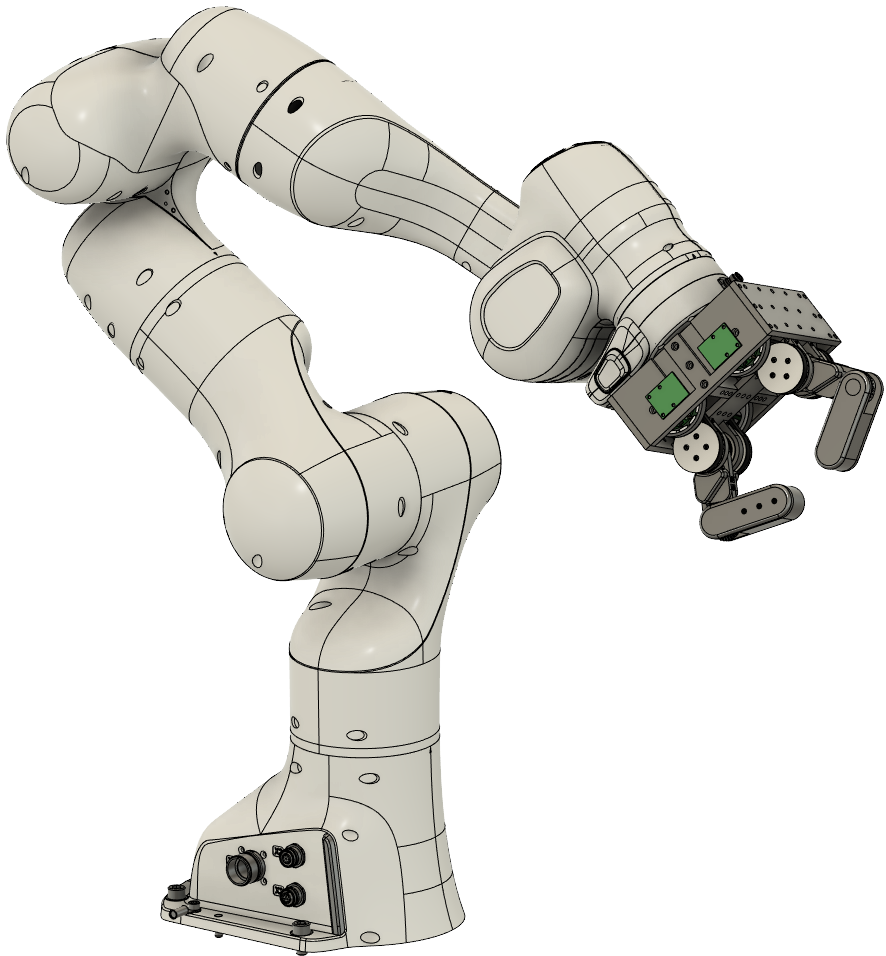}}
\\
\sidesubfloat[]{\includegraphics[width=5 cm]{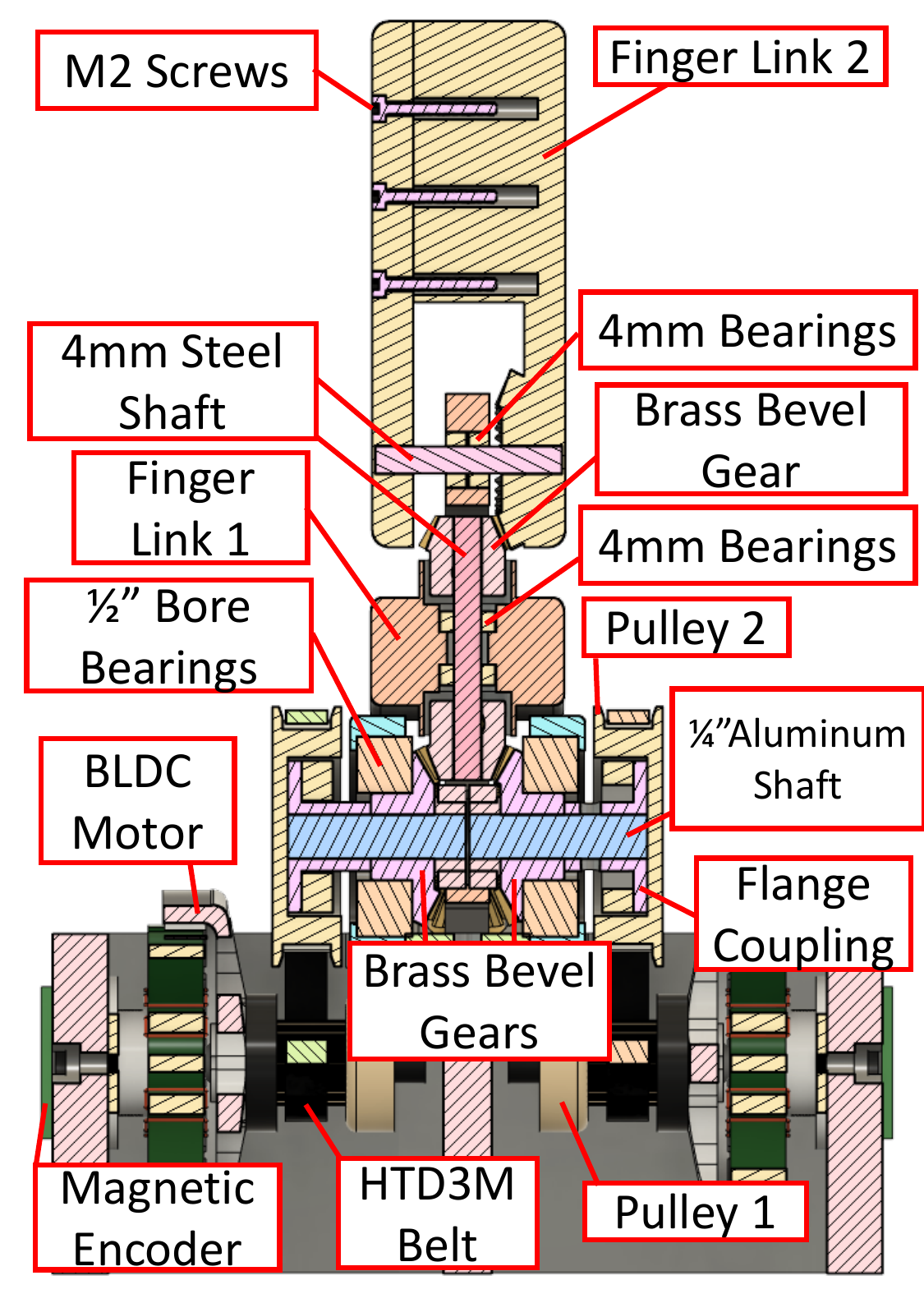}}%
\caption{Quasi-direct-drive hand design drawings, (a) Hand CAD assembly, (b) CAD rendering of the hand mounted to the Franka Emika Panda robot arm \cite{franka2}, (c) Labeled section view of the finger.}
\label{fig:HandDrawings}
\end{figure}

Table~\ref{table:design specs} presents some of the pertinent design specifications of the hand. Fig.~\ref{fig:HandDrawings} shows the completed hand CAD assembly and a labeled section view of the finger, including all of the internal components of the mechanism such as the bearings, metal shafts, belts, and pulleys.

\begin{table}[htbp]
\caption{Design Specifications of the Proposed Robotic Hand.}
\label{demo-table}
\centering
\begin{tabular}{ |c|c| } 
\hline
 \textbf{Design Parameters} & \textbf{Value} \\
\hline
 Force Output at Fingertip &  8.2 N      \\
 Motor Rating           & 360 KV (RPM/volt)   \\
 Degrees of Freedom        & 4       \\
 Operating Voltage         & 12 V       \\
 Encoder Resolution        & 16384 CPR      \\
 Hand Base Dimensions  &  133 mm $\times$ 120 mm     \\
 Finger Length   &  105.31 mm     \\
 Weight                    &  1250 g      \\
 Total Cost                &  $\sim$\$400       \\
 \hline
\end{tabular}
\label{table:design specs}
\end{table}

\subsection{Electrical Hardware and Control Architecture}
\begin{figure}[!htbp]
\centering
\includegraphics[scale=0.6]{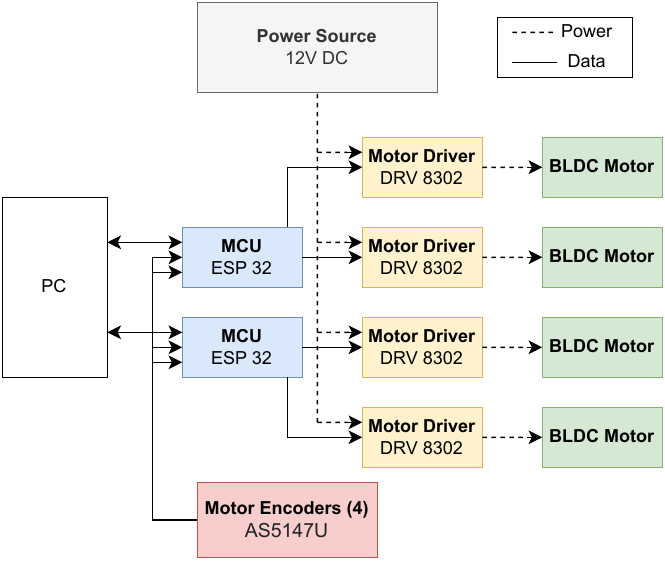}
\caption{Control architecture diagram. }
\label{fig:ControlArchitectureDiagram}
\end{figure}

The two BLDC motors (BDUAV 5010) for each finger are controlled by a single ESP 32 microcontroller as shown in Fig.~\ref{fig:ControlArchitectureDiagram}. The microcontroller is connected to a DRV8302 motor driver module, each of which is connected to a 12V DC power source and each respective motor. Magnetic motor encoders send motor position data to each motor MCU which communicates with an additional central controller via UART communication.

\section{Experiments and Results}
In this section, we present the experimental results
of the implementation of the impedance controllers and the performance of the proposed quasi-direct-drive robotic hand in grasping and manipulation of objects with unknown sizes and its interaction with an unstructured environment. The experiments are demonstrated in the attached video accompanying the paper\footnote{Video Link: \href{https://youtu.be/3bNhUlwFXkI}{https://youtu.be/3bNhUlwFXkI}}.

\subsection{Fingertip Force Output}
\begin{figure}[!htbp]
\centering
\subfloat[]{\includegraphics[scale=0.13]{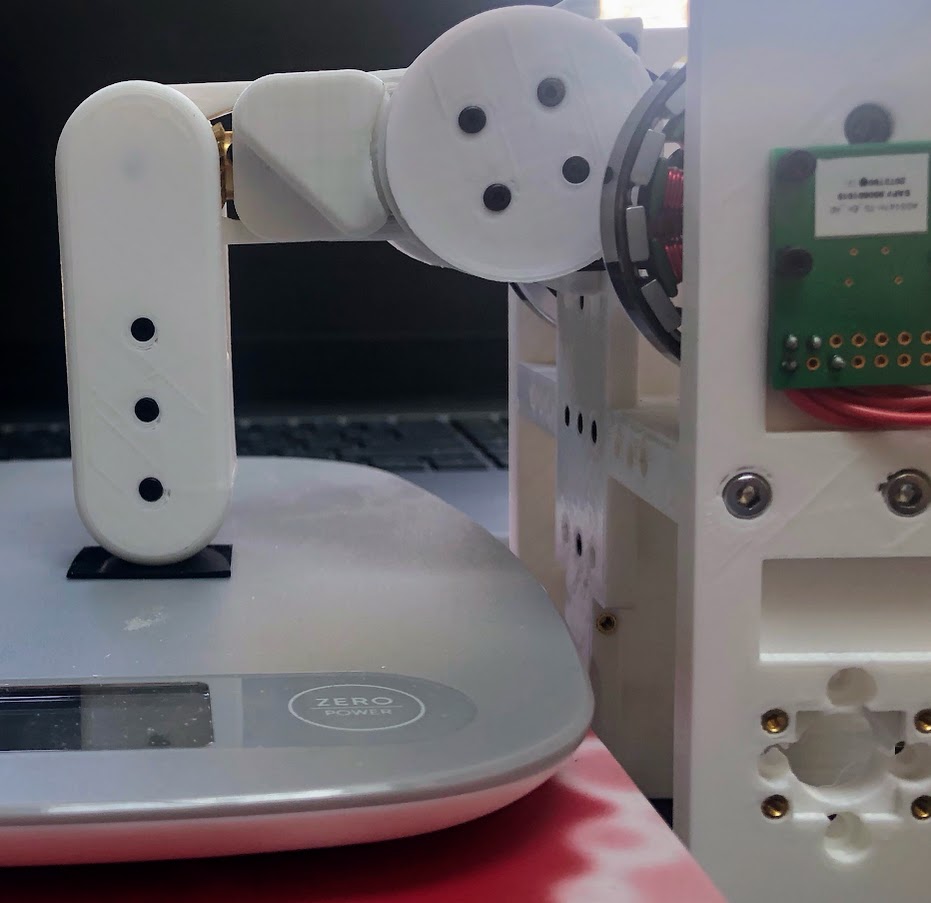}}\\
\subfloat[]{\includegraphics[scale=0.25]{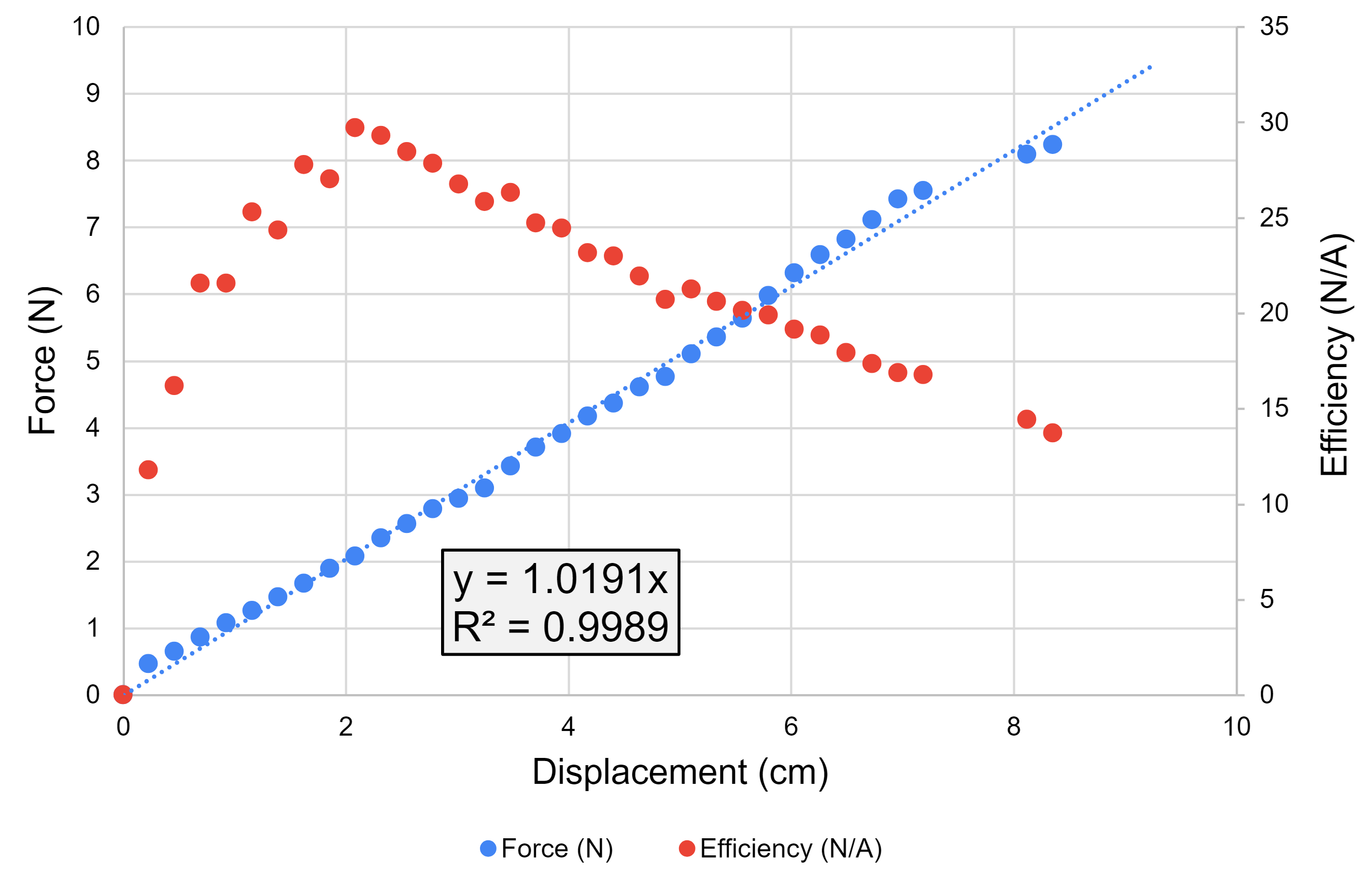}}
\caption{Measurement of force at the tip of an impedance-controlled finger, (a) Test setup, (b) Plots of fingertip force (blue) and efficiency (red) with respect to displacement.
}
\label{fig:Plot1}
\end{figure}

In order to successfully implement the impedance controller to grasp objects, the force output at the fingertip of the hand must be robustly controlled. In our case, this is done by controlling the motors torque through FOC-based torque control.
A test setup to validate the implementation and performance of the controller and a plot of force at the fingertip with respect to displacement are shown in Fig.~\ref{fig:Plot1}.
In this setup, the fingertip presses vertically on a scale that measures the force output. Displacement is measured in centimeters as the difference between the actual and desired positions in the direction orthogonal to the scale. The slope of this graph is expected to be equal to the stiffness coefficient of the impedance controller. For this test, we used a stiffness coefficient of 1 N/cm in the vertical direction which was very close to the experimental slope of 1.02 N/cm in Fig.~\ref{fig:Plot1}-b.


The efficiency of the force output is also shown in Fig.~\ref{fig:Plot1}-b. Here, efficiency is considered to be the force output divided by the total current draw of the system as measured by the 12V DC power supply. The motors run most efficiently where the force output is just over 2N. While not possible for all tasks, it would be beneficial to run the motors within this region of force output.

\subsection{Stable Grasps in Response to Disturbance}
\begin{figure}[!htbp]
\centering
\subfloat[]{\includegraphics[height = 3.1 cm]{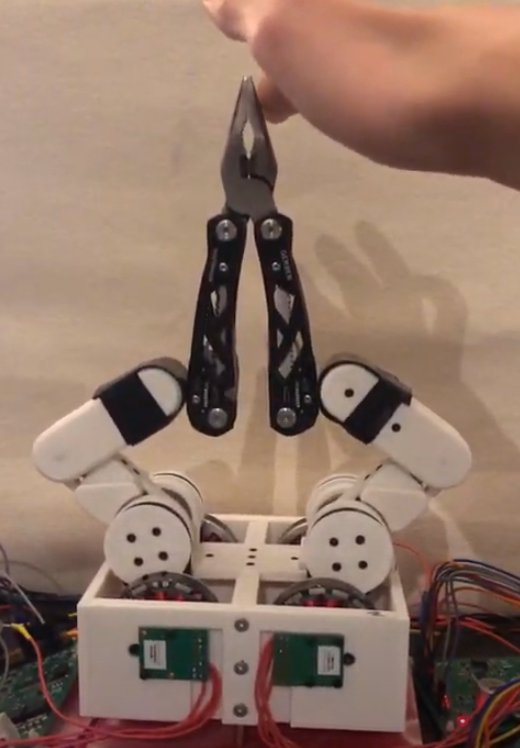}} \!
\subfloat[]{\includegraphics[height = 3.1 cm]{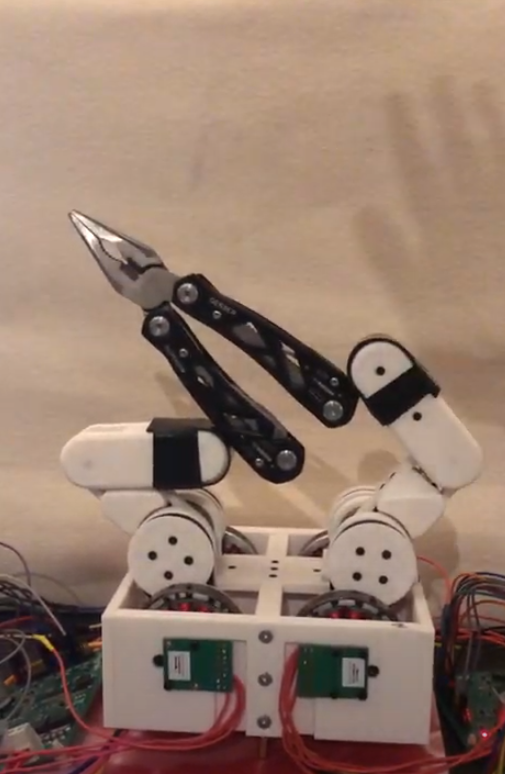}}\!
\subfloat[]{\includegraphics[height = 3.1 cm]{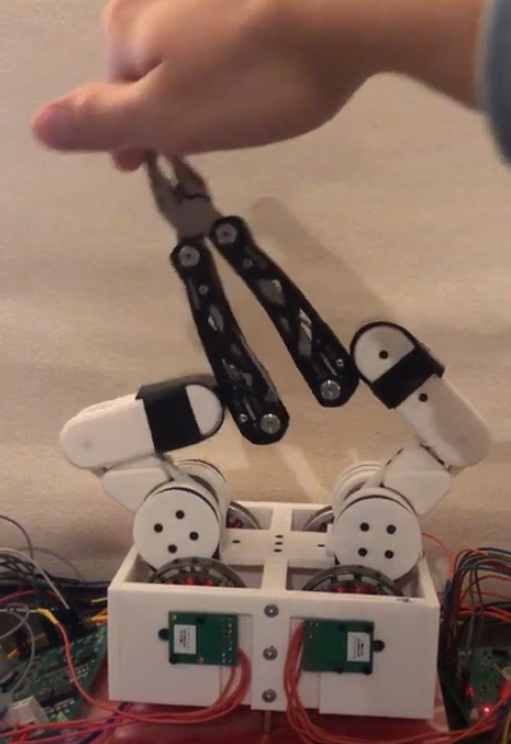}}\!
\subfloat[]{\includegraphics[height = 3.1 cm]{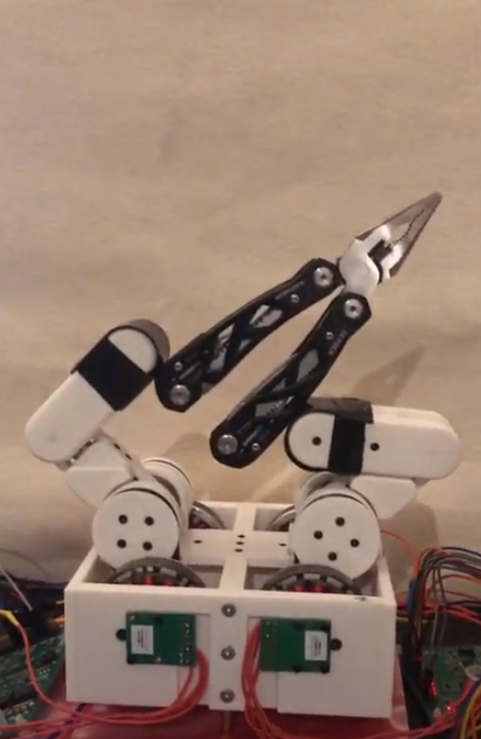}}
\caption{Stable grasp of a pair of pliers in the presence of external wrenches.}
\label{fig:pliers}
\end{figure}

Robotic hands and grippers must be able to maintain a stable grasp in response to external disturbances. Impedance control can help make this possible. When performing a force-closure grasp (Fig.~\ref{fig:pliers}), the fingers attempt to maintain a desired position, but they are also compliant, allowing movement in response to external forces in accordance with the defined stiffness and damping matrices. When an external wrench is applied to the pliers, the fingers retain rolling contact friction between the fingertips and the handles of the pliers. This allows the pliers to resist slipping from the grasp of the hand. After the external force is applied, the weight of the pliers prevents the fingers from returning to their original position, but a stable grasp is successfully maintained despite the hand having only two fingers. For these grasps, the Cartesian impedance control model is used as we are primarily concerned with applying force on the object by the fingertips in the Cartesian space.

\subsection{Form-Closure Grasp}
\begin{figure}[!htbp]
\centering
\subfloat[]{\includegraphics[height = 2.35 cm]{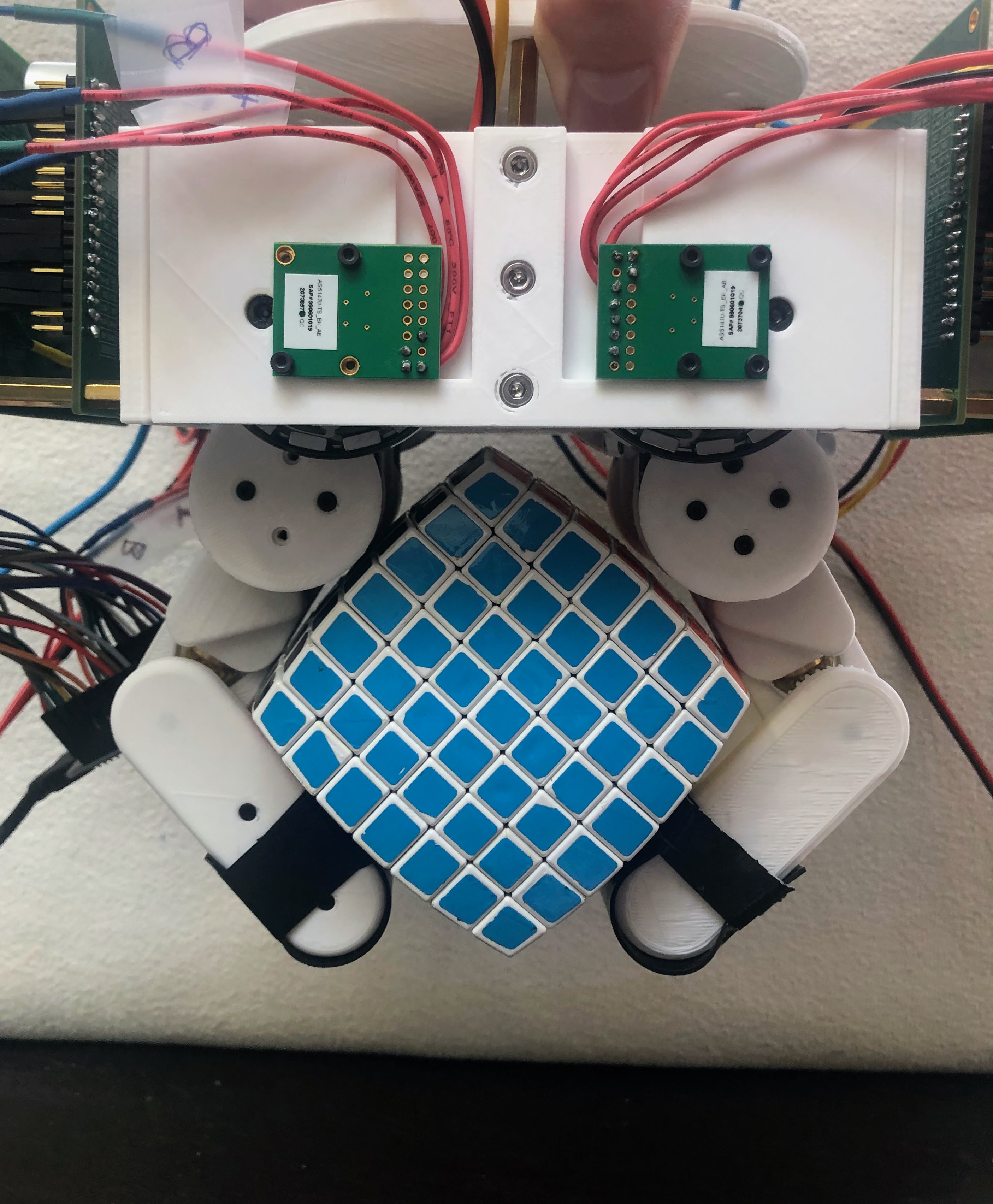}}\!
\subfloat[]{\includegraphics[height = 2.35 cm]{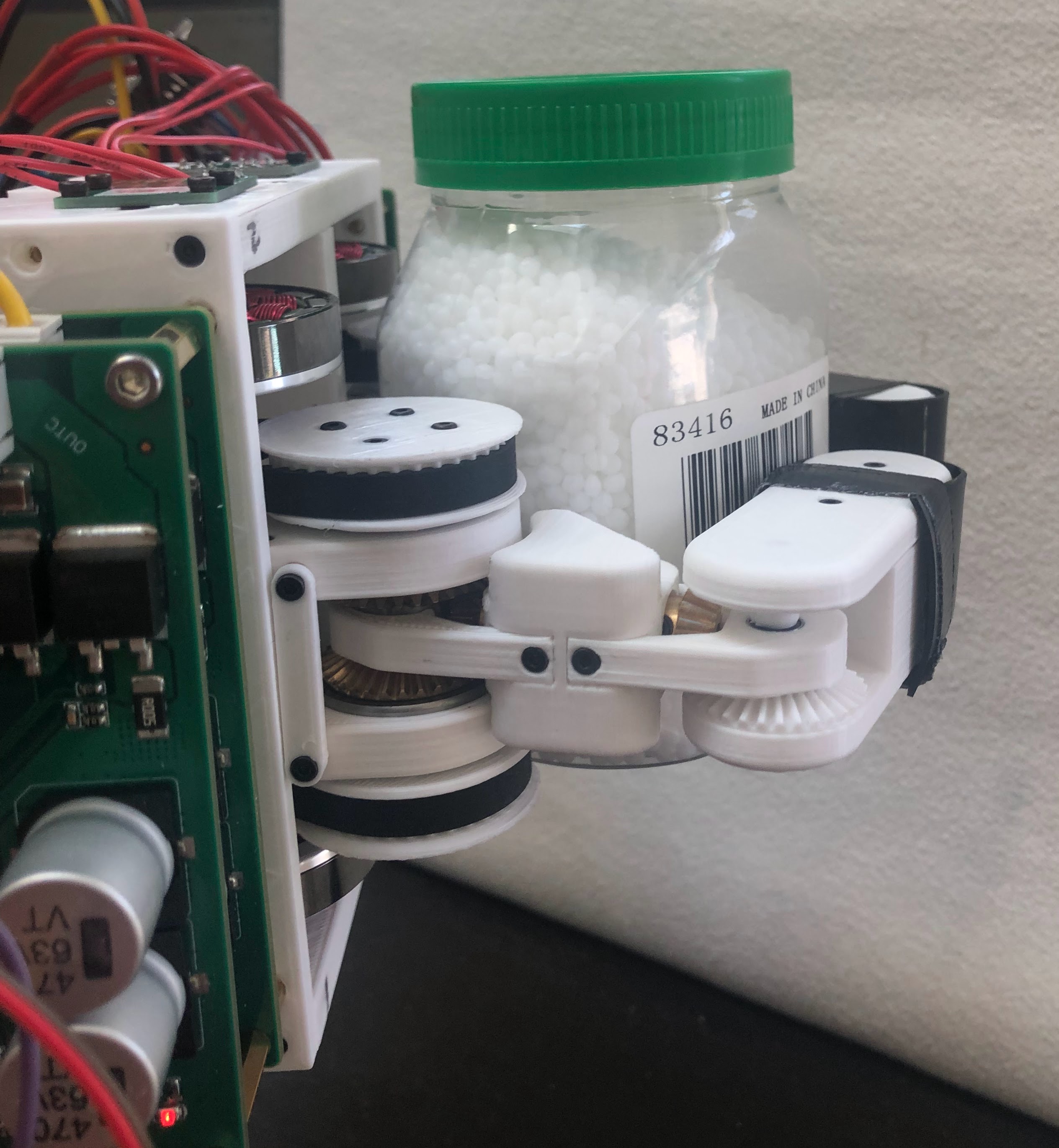}}\!
\subfloat[]{\includegraphics[height = 2.35 cm]{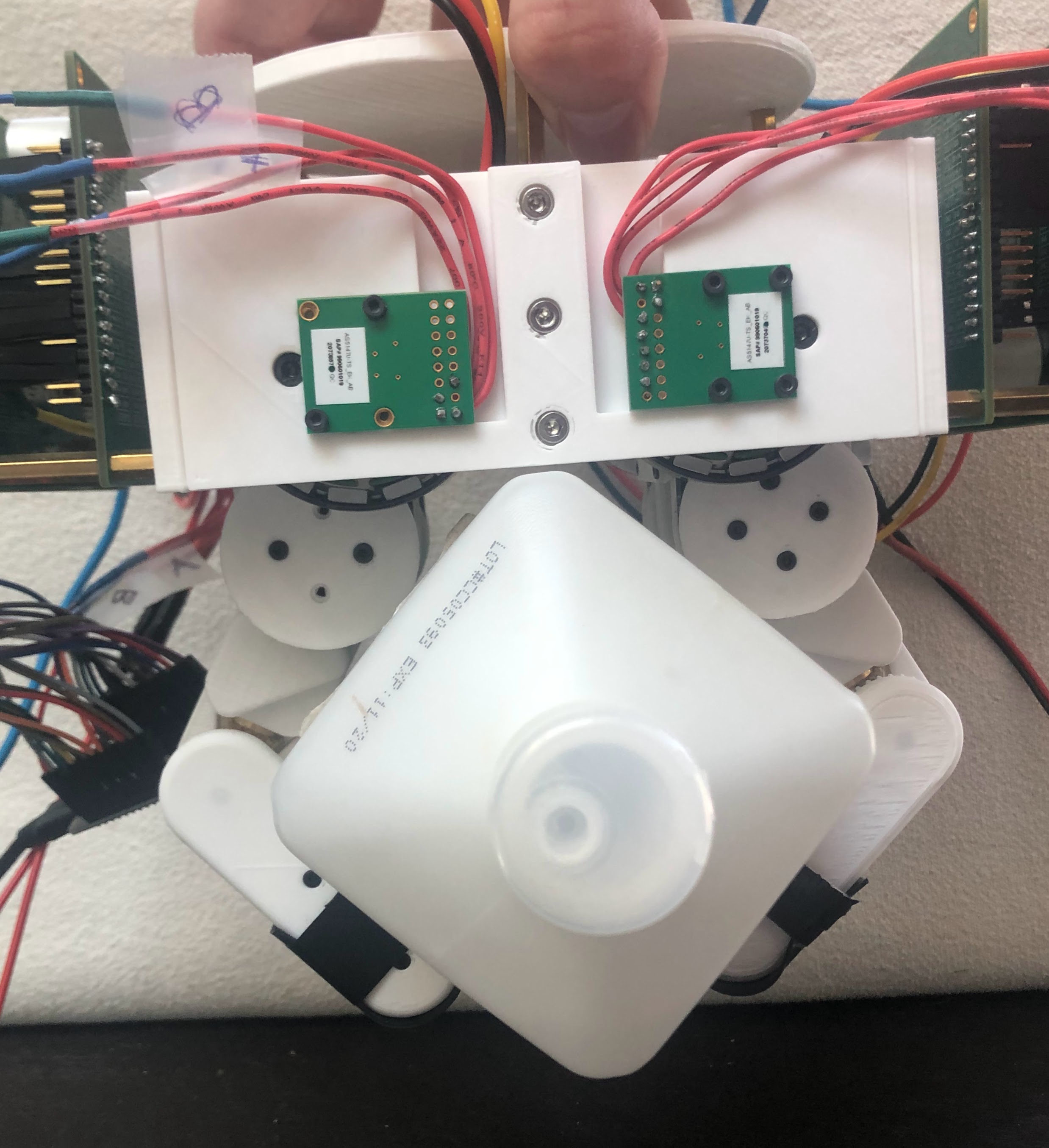}}\!
\subfloat[]{\includegraphics[height = 2.35 cm]{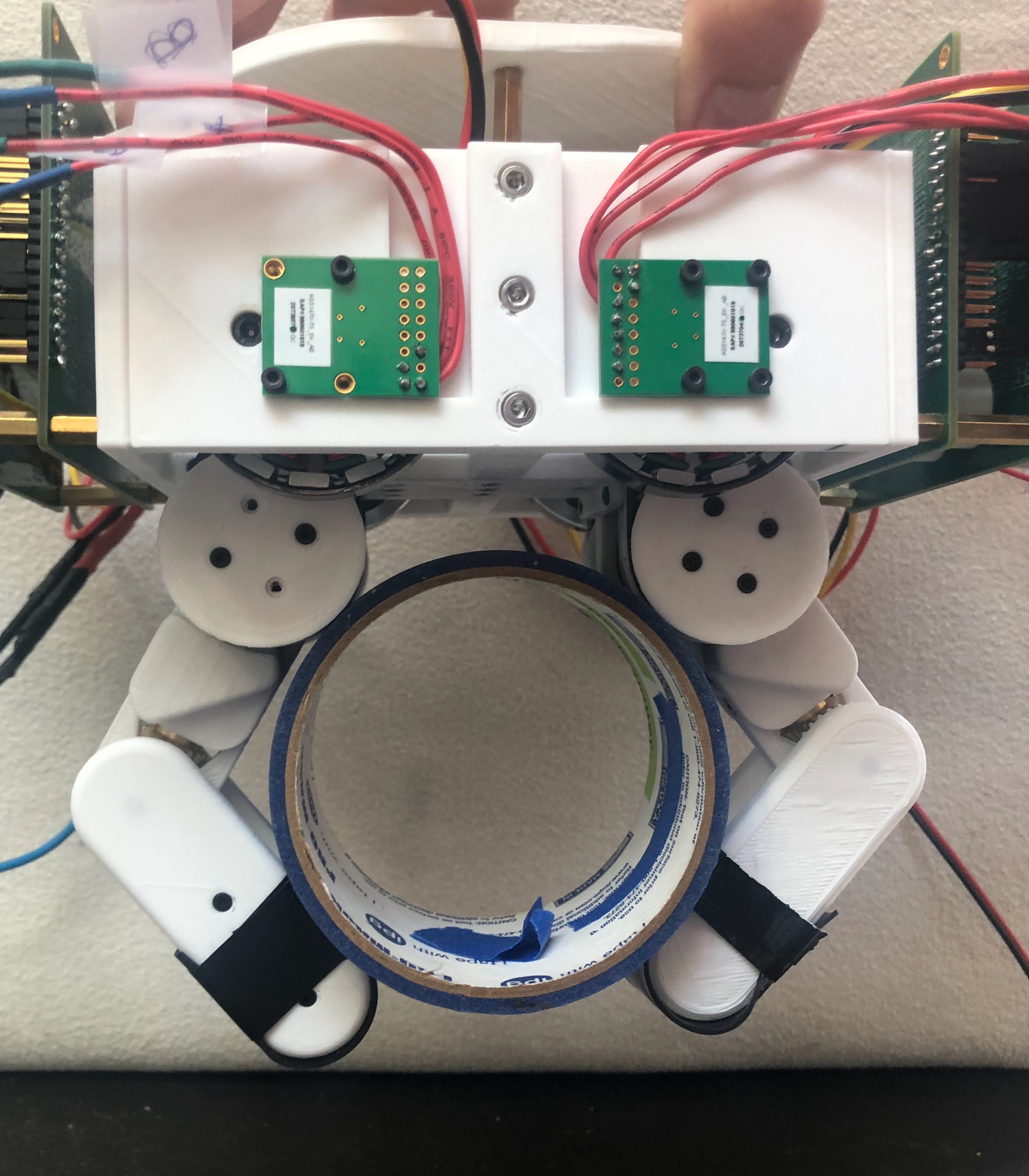}}
\caption{Form-closure grasp of various objects.}
\label{fig:form}
\end{figure}

Our hand is unique compared to other prior DD and QDD hands (e.g., \cite{Bhatia2019DirectDH,9981569,QuasiDirectDrive_Lin,TriFinger}) in its ability to perform both force-closure and form-closure grasps. In contrast to force closure, form closure achieves a stable grasp through the shape of the gripper matching the shape of the object being grasped. 
We implement the impedance control in the joint space rather than Cartesian space for form closure because the objects contact the finger along the faces of the two finger links and not just at the tips of the fingers. Using this technique, we can form the fingers around a variety of objects of different unknown sizes and shapes as shown in Fig.~\ref{fig:form}.

\subsection{Smack-and-Snatch Manipulation}
\begin{figure}[!htbp]
\centering
\includegraphics[width = \textwidth]{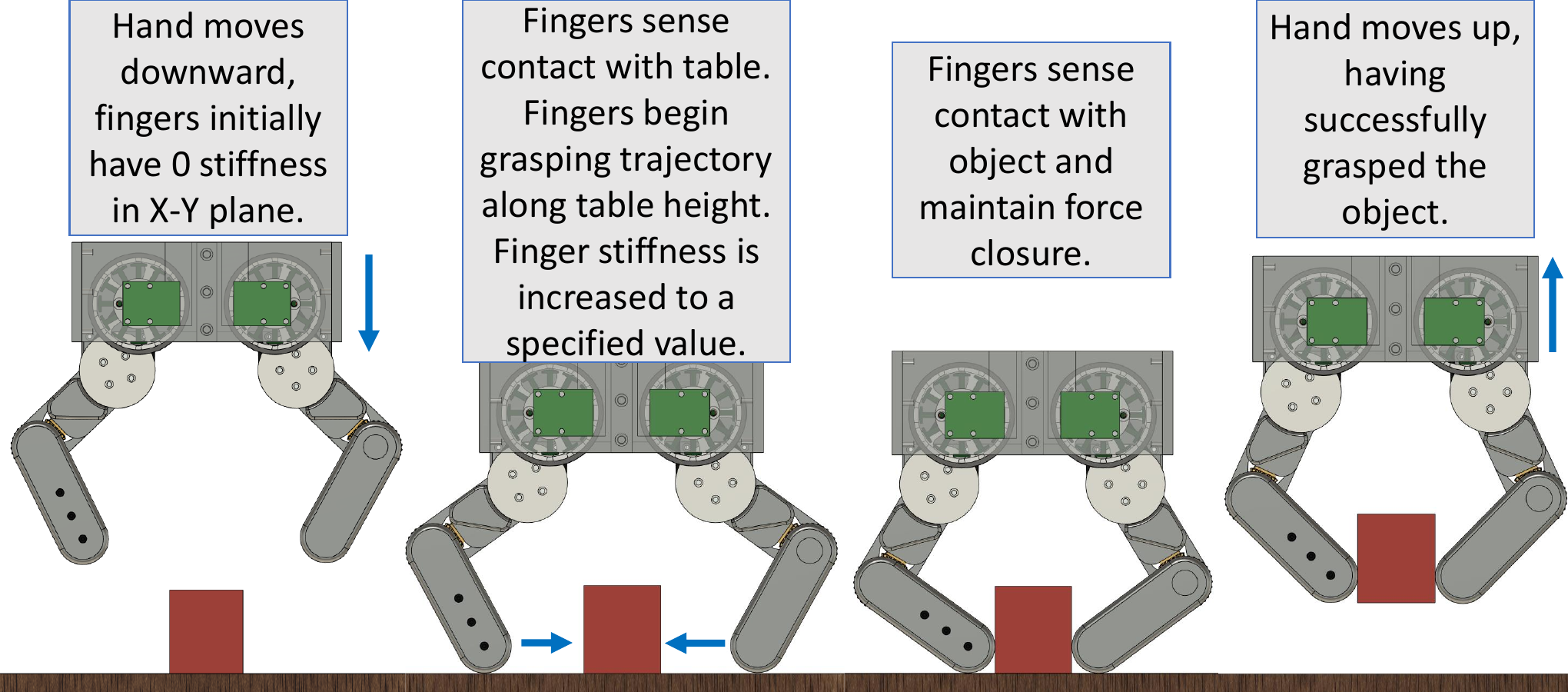}
\caption{Visualization of smack-and-snatch manipulation with our QDD robotic hand.}
\label{fig:smack and snatch}
\end{figure}

\begin{figure}[!htbp]
\centering
\subfloat[]{\includegraphics[height = 2.7 cm]{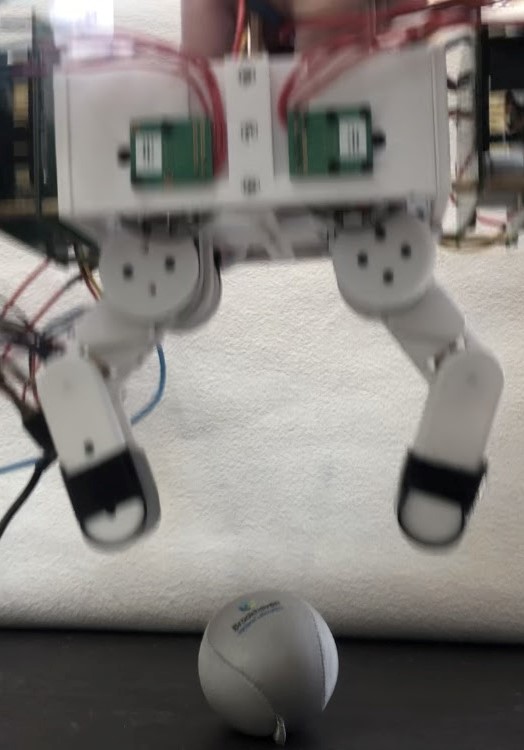}}\!
\subfloat[]{\includegraphics[height = 2.7 cm]{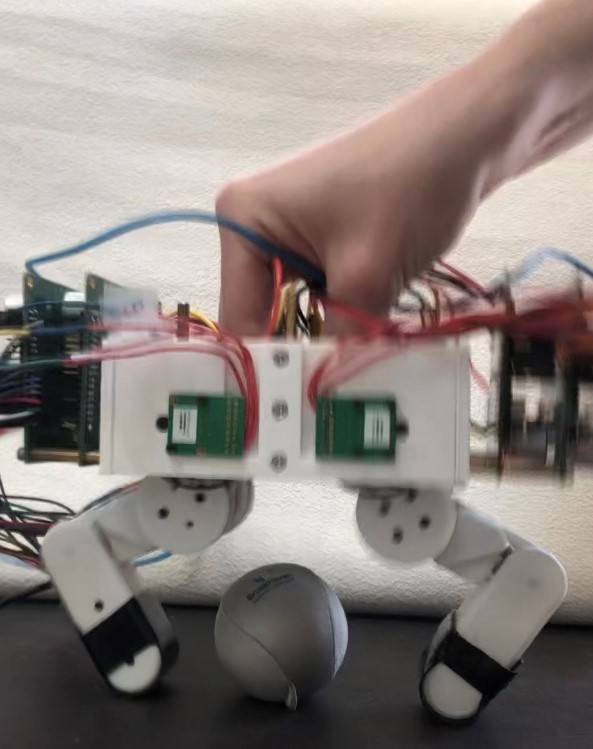}}\!
\subfloat[]{\includegraphics[height = 2.7 cm]{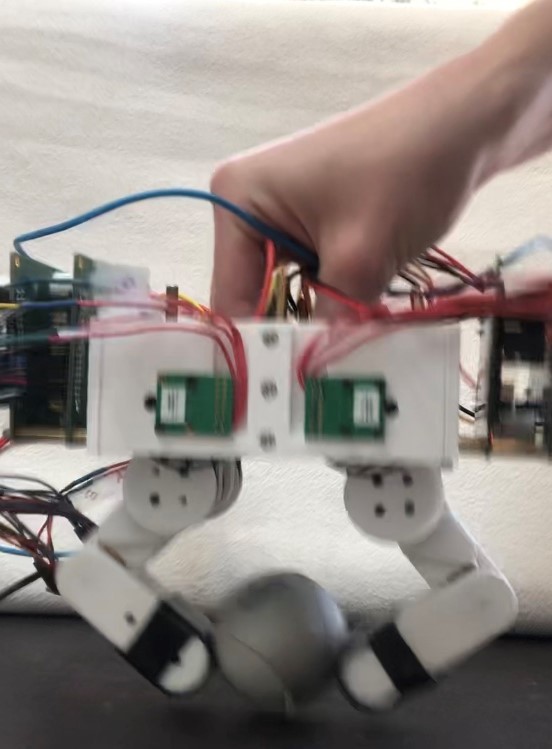}}\!
\subfloat[]{\includegraphics[height = 2.7 cm]{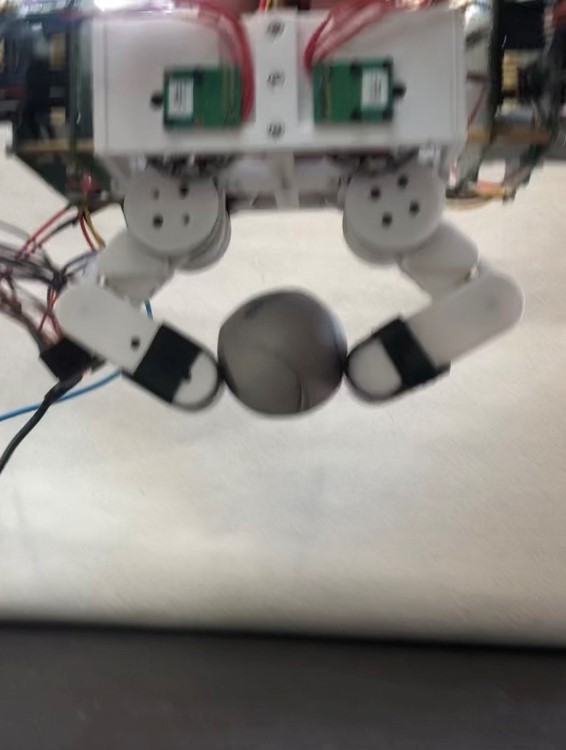}}
\caption{Smack-and-snatch manipulation of a ball.}
\label{fig:sas}
\end{figure}

\textit{Smack-and-snatch} refers to a dynamic manipulation task first coined and performed by the CMU DDHand \cite{Bhatia2019DirectDH}, and later recreated by Lin \textit{et al.} \cite{QuasiDirectDrive_Lin} with their linkage based QDD robotic hand. The sequence of smack-and-snatch manipulation is visualized with our QDD hand in Fig.~\ref{fig:smack and snatch}. In this type of manipulation, the robotic hand/gripper must grasp an object from overhead at an unknown height. Since these hands have transparent, backdriveable drive-trains, the hand can safely sense contact with the environment using finger position sensors, and then, close the fingers of the hand after sensing contact with the environment. Because of the low inertia and friction of the fingers, the impact imparted on the hand by the environment is low, and the hand does not get damaged even at high velocities. This compliance in the hand allows the fingers to act as a contact sensor with the environment, and trigger the grasping trajectory of the hand. This method demonstrates how the fingers of our QDD hand can be used as a sensor to sense contact with the environment which can be exploited during a task. In addition, this design allows for potentially highly dynamic grasping at high speeds because the manipulator does not need to pause for a successful grasp. Since the hand has not been mounted to the Franka Emika Panda robot arm yet, we tested this concept by simply moving the robotic hand manually up and down in a fast, vertical trajectory as shown in Fig.~\ref{fig:sas}. Initially, the fingers each have zero stiffness. When the fingers are displaced by a specified value in the vertical direction by the environment, the stiffness matrix changes to some constant in the $X$ and $Y$ directions. At this point, the fingers each travel to their desired positions located in the center of the hand's workspace. This enables the hand to grasp the ball of unknown size in a quick, smooth motion without the need to pause at the bottom of the hand trajectory, and without the need for extra sensors to locate the table position.
This method can be also applied to the grasping of fragile objects (e.g., eggs).

\subsection{In-Hand Manipulation: Object Rotation and Translation}
\begin{figure}[!htbp]
\centering
\subfloat[]{\includegraphics[height = 2.2 cm]{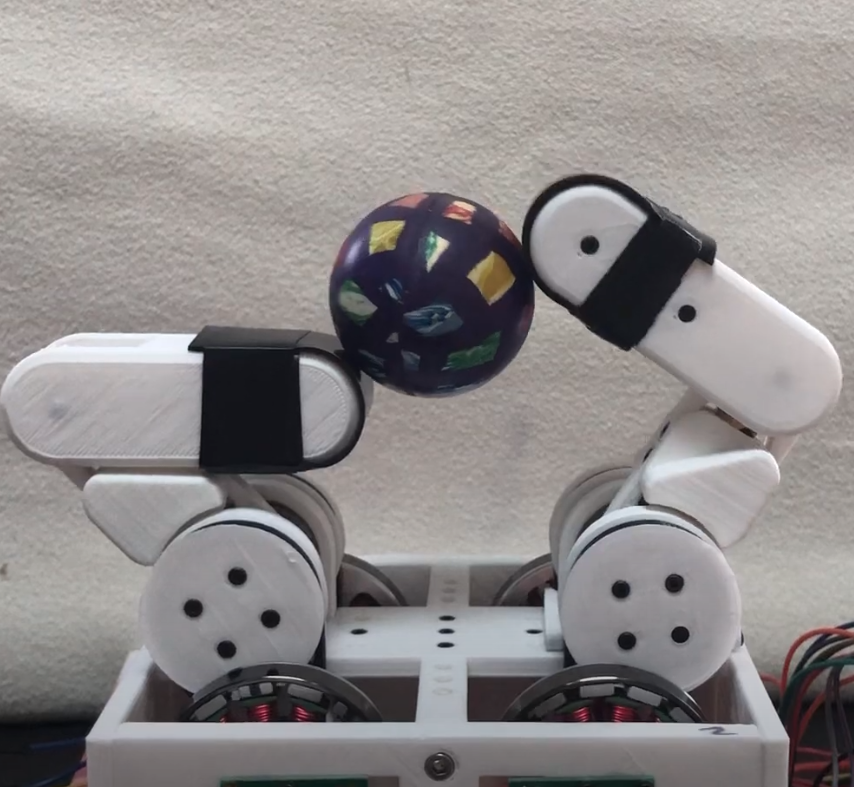}}\!
\subfloat[]{\includegraphics[height = 2.2 cm]{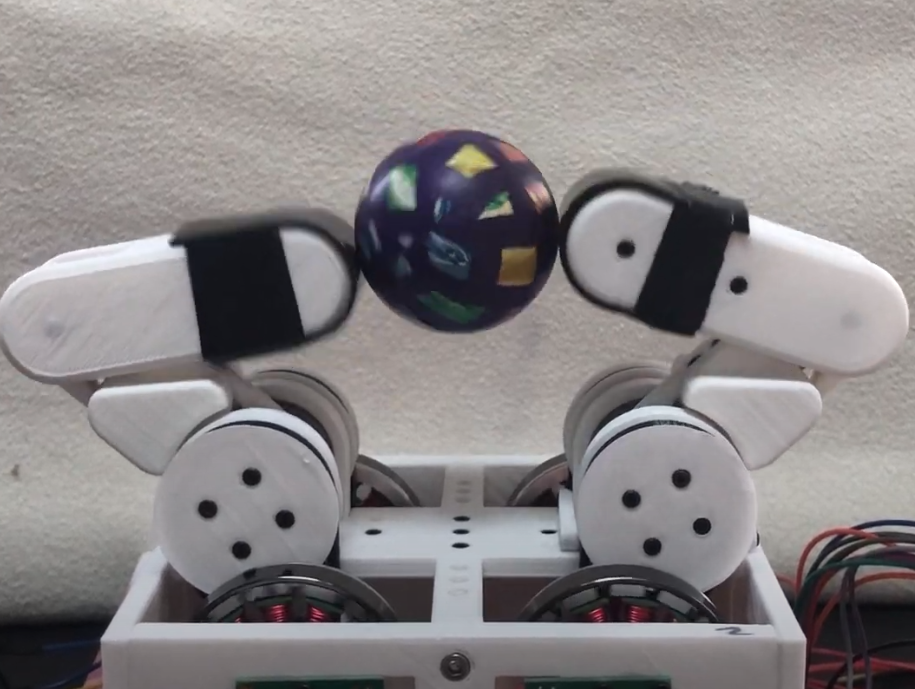}}\!
\subfloat[]{\includegraphics[height = 2.2 cm]{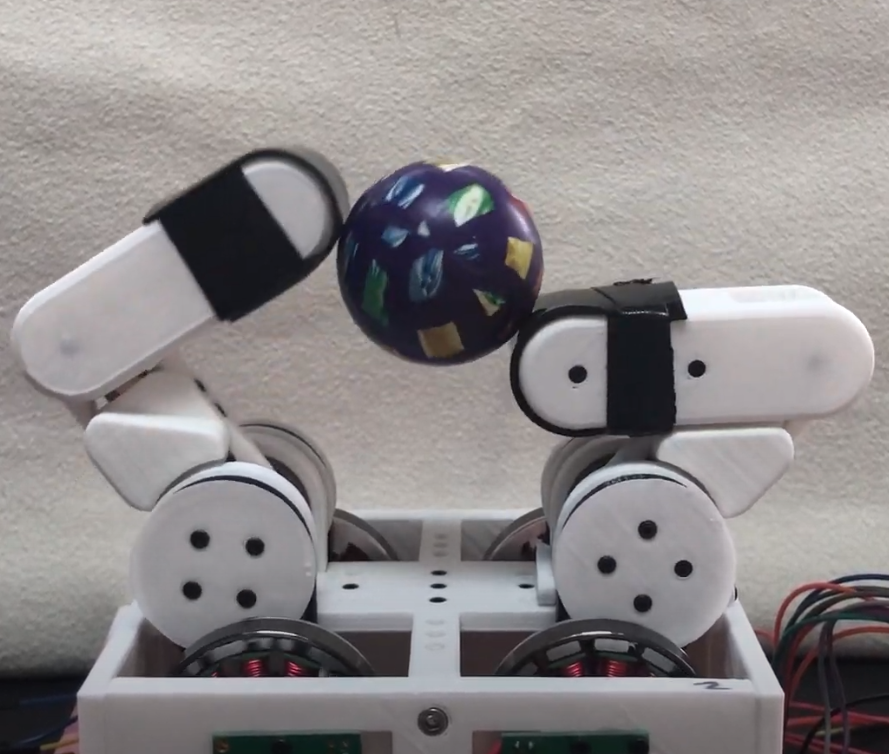}}\!
\caption{In-hand manipulation of a rubber ball.}
\label{fig:ball}
\end{figure}

In-hand manipulation is defined by the motion of the object in the robotic hand after it has been grasped. Due to the geometry of our hand and the impedance controller used, the fingers can maintain contact force on the object while the fingers move through a specified trajectory. A sample task is demonstrated in Fig.~\ref{fig:ball}. The fingers of the hand rotate a ball while moving it up and down. The fingers have a linear desired trajectory in the $Y$-direction while trying to maintain a constant position in the $X$-direction less than that of the diameter of the rubber ball. The difference between the desired and actual finger positions in the $X$-direction is the source of the gripping force on the ball. Since Cartesian impedance control is being used here, the force applied is proportional to the stiffness coefficient in the $X$-direction. As shown, the fingers can maintain rolling contact with the object as it is manipulated by the fingers. Rolling contact introduces nonholonomic constraints on the system which can be exploited to manipulate the object \cite{cui_sun_dai_2017}. This allows the hand to both rotate and translate the grasped object and greatly expands the capabilities of the hand compared to hands/gripers that can only pick and place objects.

\subsection{In-Hand Manipulation: Object Regrasping}
\begin{figure}[!htbp]
\centering
\subfloat[]{\includegraphics[height = 2.1 cm]{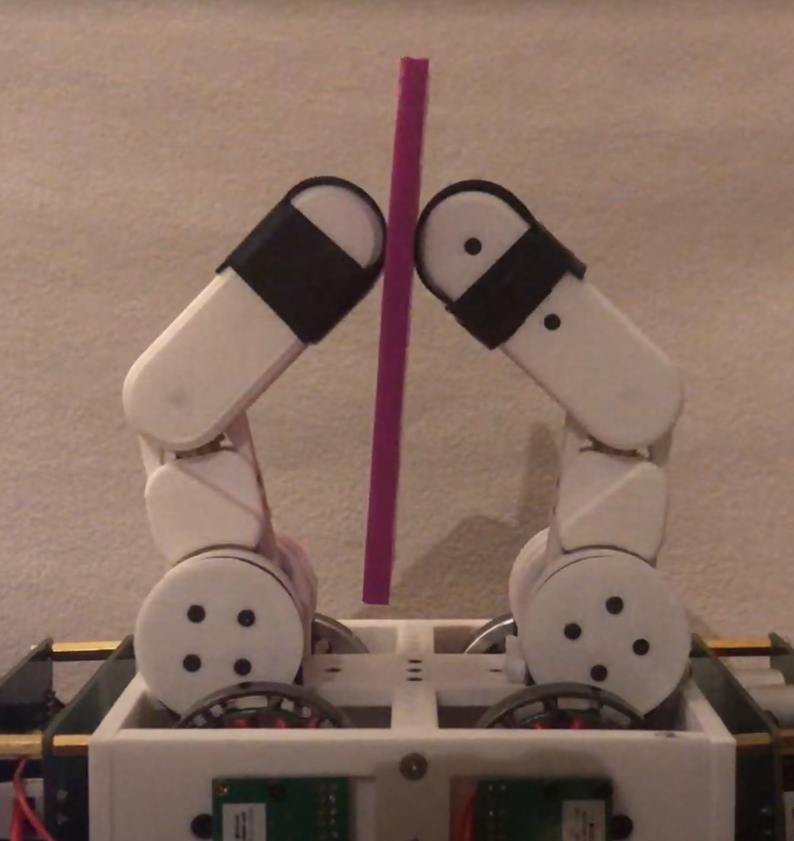}}\!
\subfloat[]{\includegraphics[height = 2.1 cm]{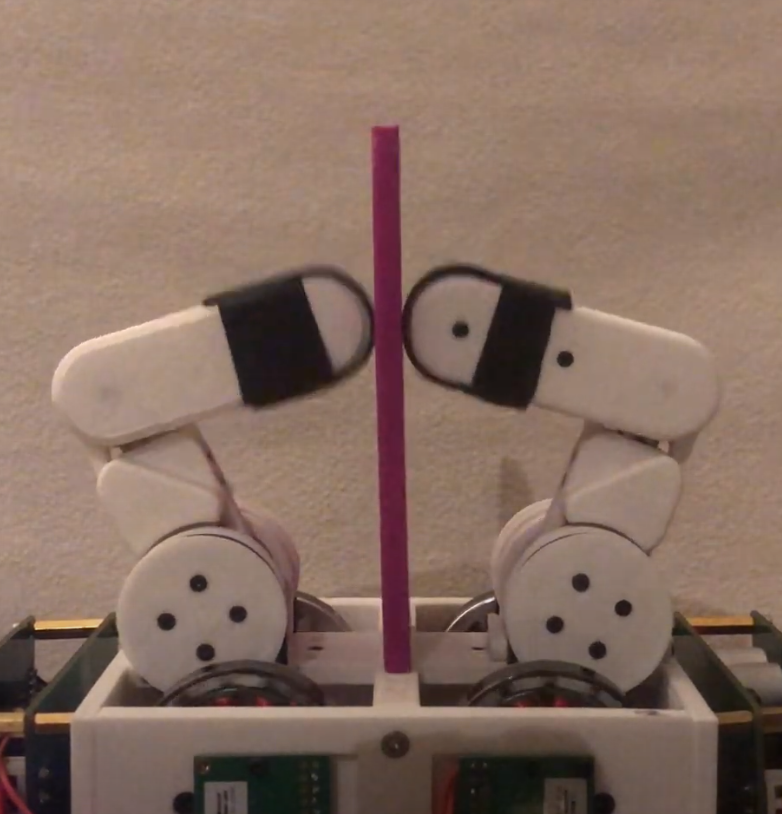}}\!
\subfloat[]{\includegraphics[height = 2.1 cm]{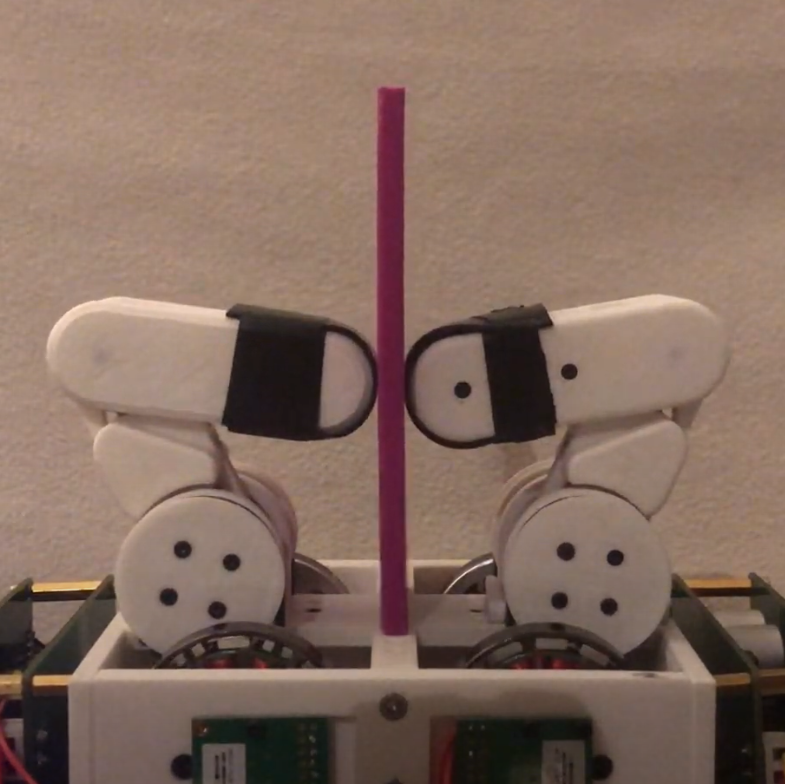}}\!
\subfloat[]{\includegraphics[height = 2.1 cm]{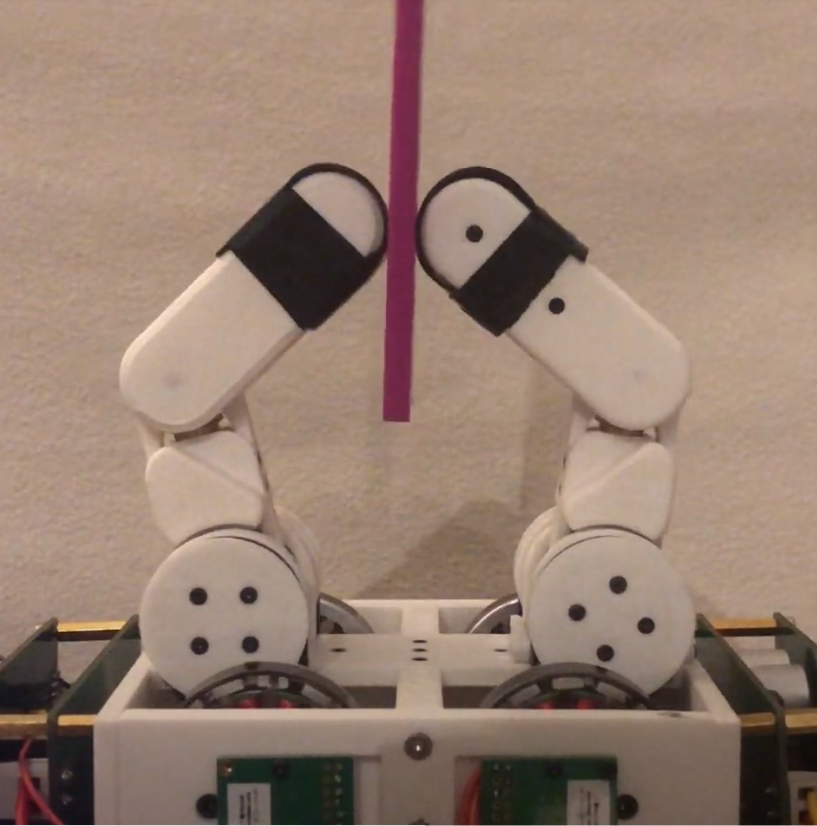}}
\caption{In-hand manipulation and object regrasping through pushing.}
\label{fig:push}
\end{figure}

Several studies demonstrated in-hand manipulation using basic grippers by leveraging interactions with the environment, a concept referred to as ``extrinsic dexterity" \cite{6907062}. For example, Chavan-Dafle \textit{et al.} \cite{7354264} model the contact forces on a grasped object when it is manipulated by pushing it against its environment. Using this method, the object can be re-positioned in the two fingers of the gripper, however, since the gripper only has 1 DOF, it must be connected to a robot arm. Using our hand, we can perform a similar operation, but instead of using contact with the environment, we can exploit contact with the palm of the hand. When grasping an object with the fingertips and moving the fingertips in a trajectory toward the palm, the palm applies a force to the object, pushing the object up in relation to the fingertips as shown in Fig.~\ref{fig:push}. To facilitate the sliding contact between the fingertip and the object, the stiffness coefficient is reduced by half in the direction normal to the surface of the object ($X$-direction) while the fingers push the object against the palm. This is simply a heuristic chosen to reduce the frictional force to make sliding easier, while still maintaining a stable grasp. Using gravity, the fingers can reset the object's original position in the hand and the task can be repeated with great reliability.

\subsection{Picking up a Coin}
\begin{figure}[!htbp]
\centering
\includegraphics[width = 4 cm]{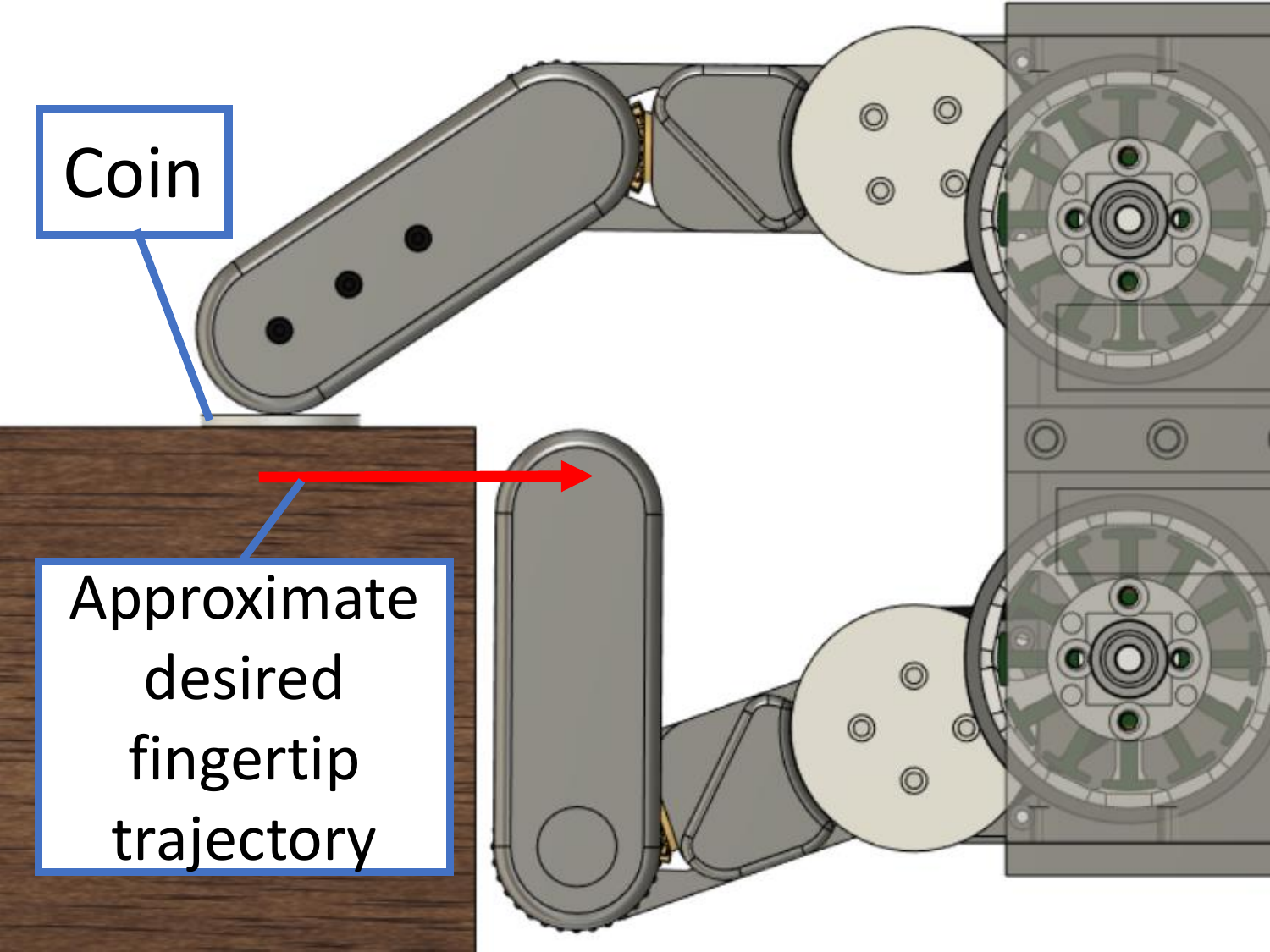}
\caption{Configuration of the robotic hand for picking up the coin. }
\label{fig:coin traj}
\end{figure}

\begin{figure}[!htbp]
\centering
\subfloat[]{\includegraphics[height = 2.1 cm]{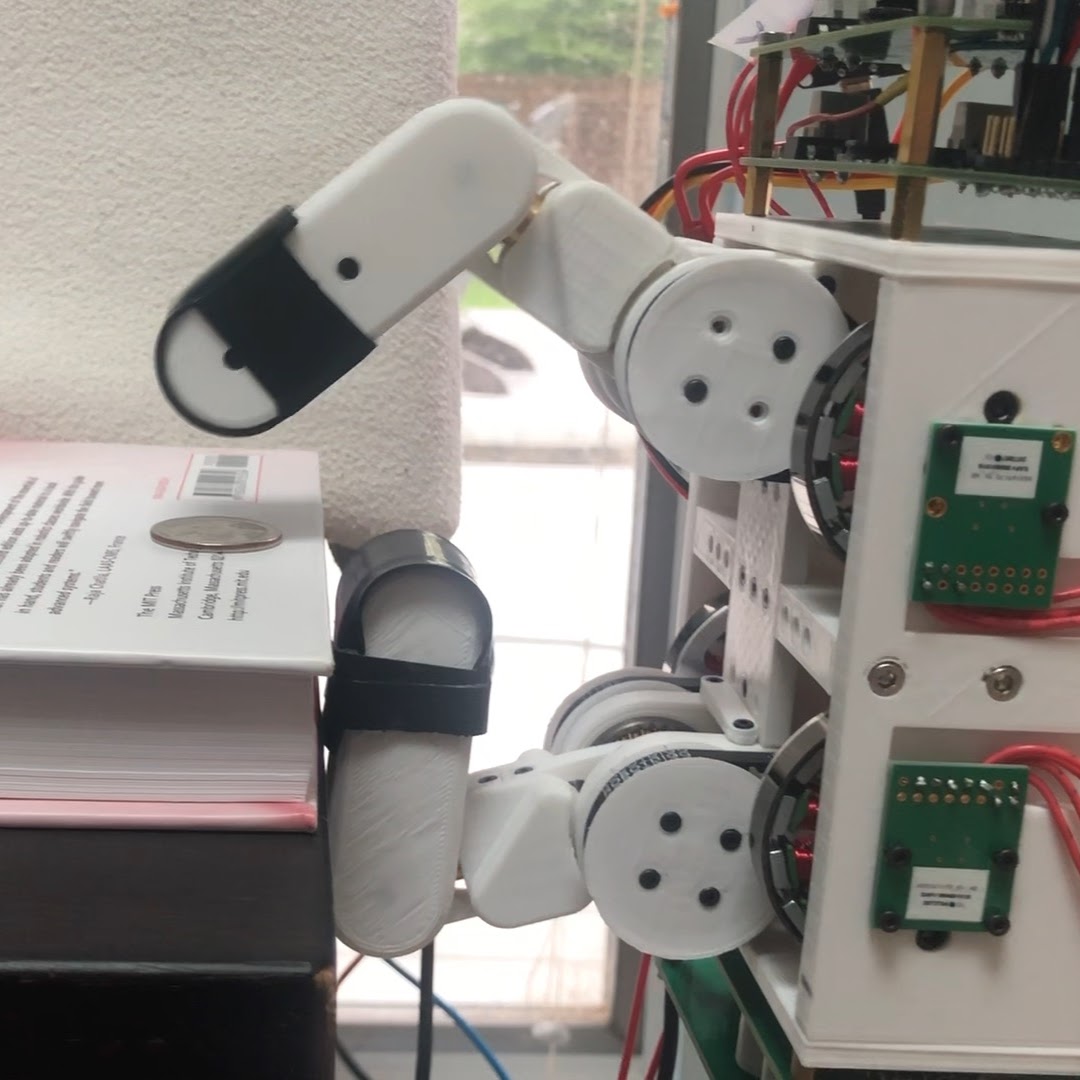}}\!
\subfloat[]{\includegraphics[height = 2.1 cm]{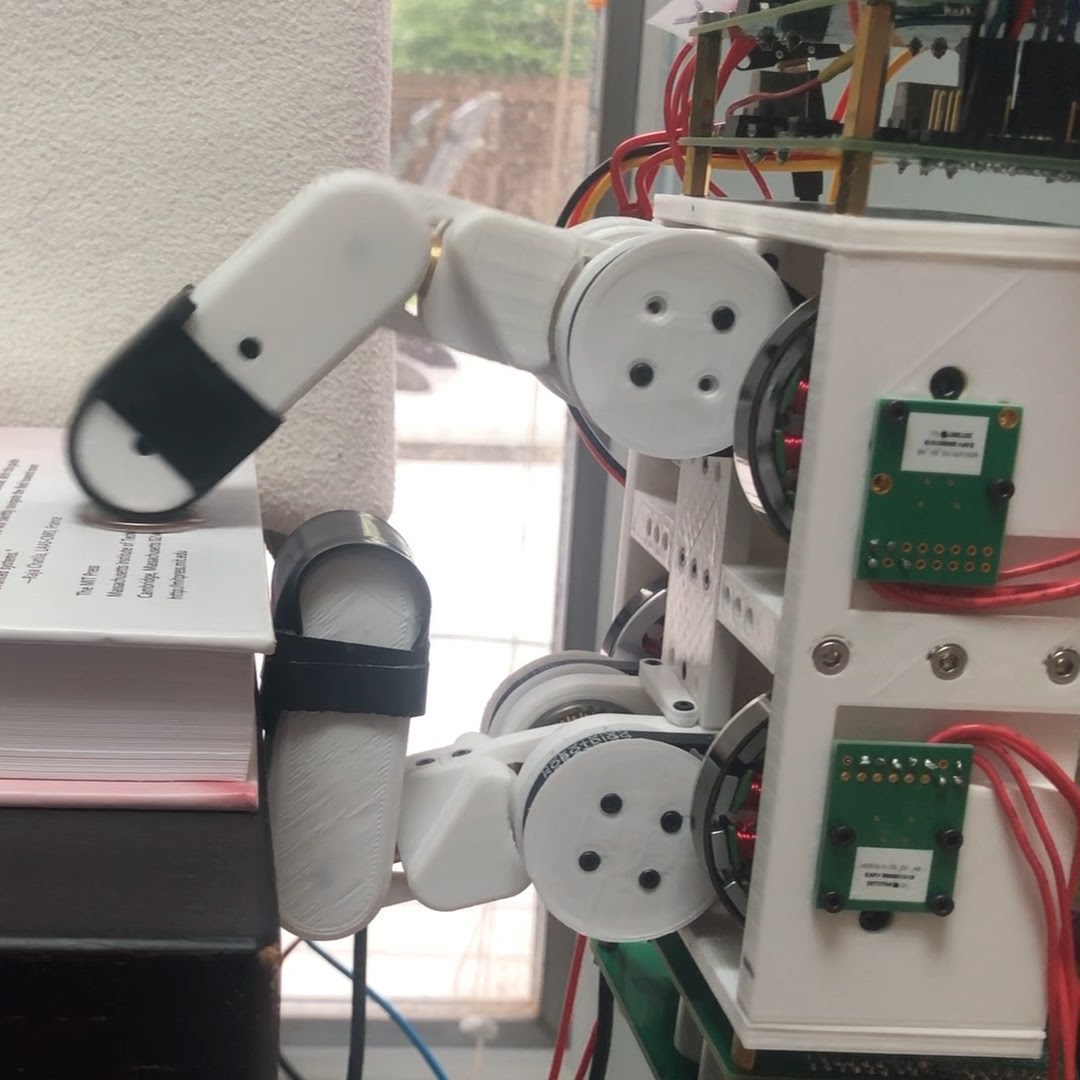}}\!
\subfloat[]{\includegraphics[height = 2.1 cm]{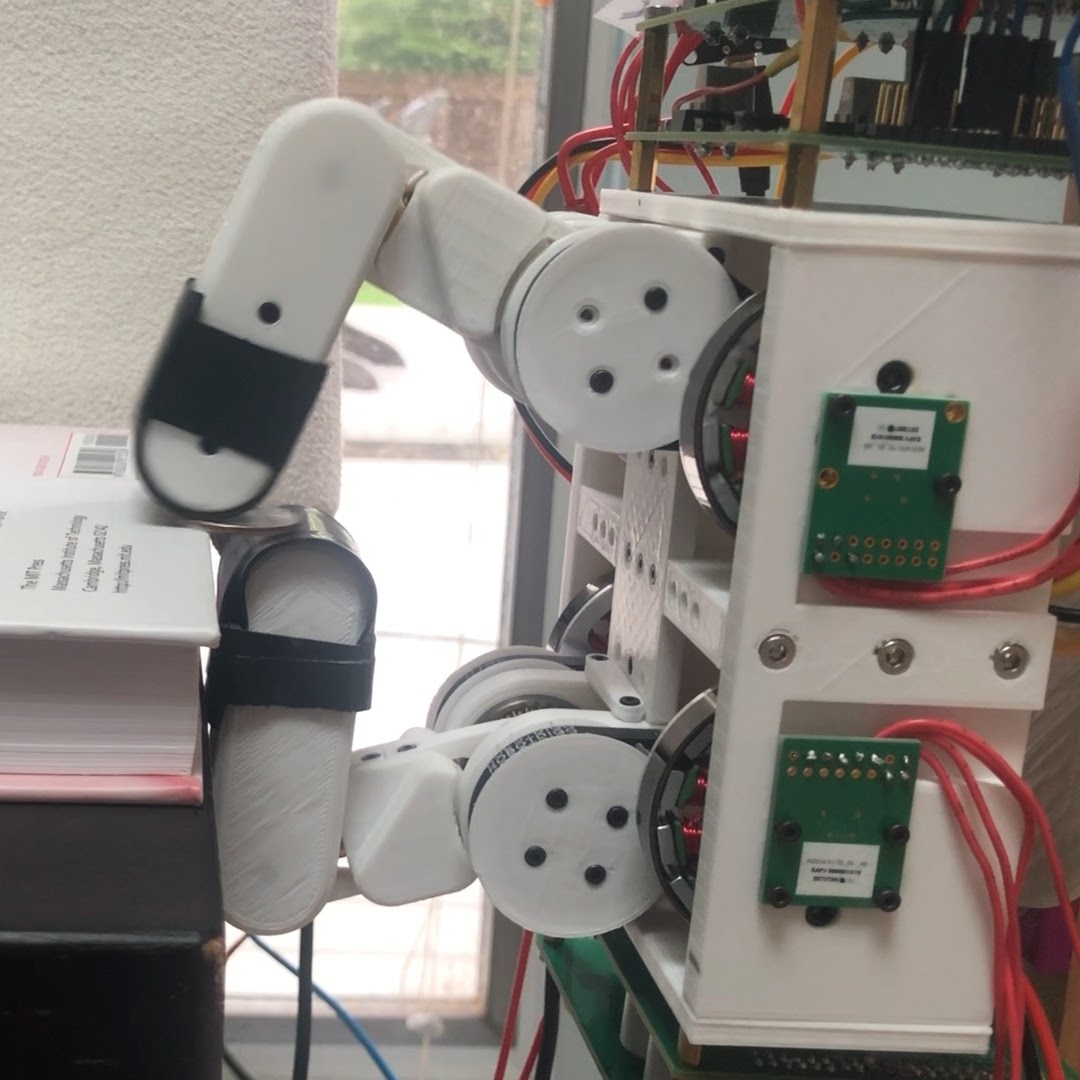}}\!
\subfloat[]{\includegraphics[height = 2.1 cm]{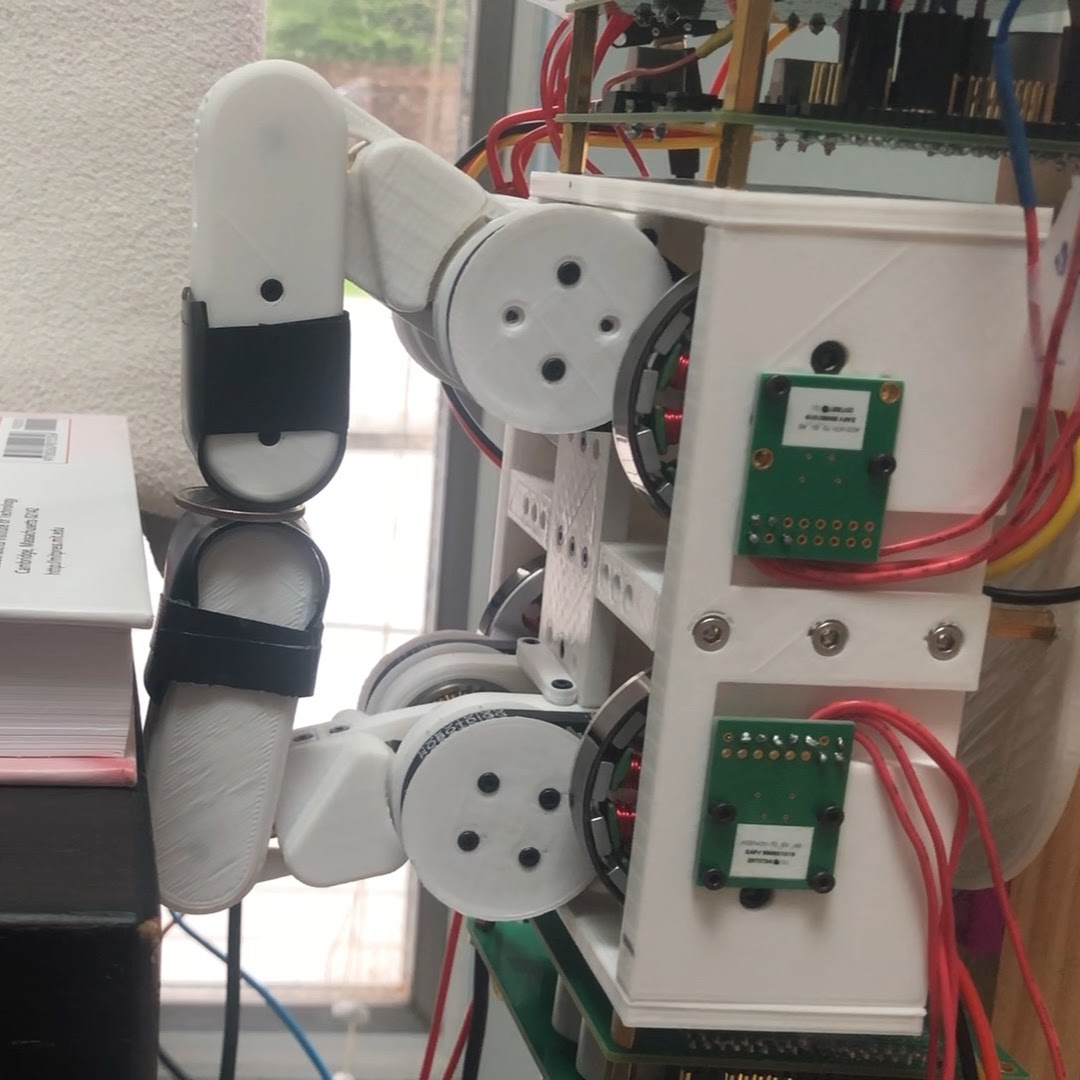}}
\caption{Picking up a coin.}
\label{fig:coin}
\end{figure}

One task robot hands and grippers may struggle with is picking up small flat objects such as a coin from a surface. Due to the geometry of the fingers in our robot hand, it would be very difficult to lift a coin directly from the surface of a table. However, if we position the hand near the edge of the table (Fig.~\ref{fig:coin traj}), one of the fingers can slide the coin along the surface towards the edge of the table, and into a stable grasp between the fingertips, by applying a force normal to the coin, while the other finger can safely maintain contact with the table without knowing the exact position/geometry of the edge of the table (Fig.~\ref{fig:coin}).
This normal force remains constant by maintaining the difference between the actual and desired positions in the direction normal to the table, in our case the $X$ direction.

\section{Conclusion and Future Work}
When robotic hands and grippers are deployed in the real world, they are likely to undergo unplanned contact with the environment.
We believe that this work shows some of the benefits of using quasi-direct-drive actuators with impedance control for applications in general-purpose robotic hands and grippers. A transparent, backdrivable gear reduction effectively turns the motor into an additional force sensor which can aid in dexterous manipulation and highly dynamic tasks, while field-oriented control is implemented to accurately control motor torque with current sensing and two PI controllers. We define transparency as the ease to which force and motion can be applied not only from the finger mechanism to the environment but also from the environment to the finger and actuator. This is essential for tasks such as \textit{Smack-and-snatch} manipulation, and grasping delicate objects with a light touch. Impedance control proves to be a simple and intuitive way of adjusting motor/joint angular position and compliance.
Unlike series-elastic and sensored approaches, quasi-direct-drive requires only current sensing enabling higher bandwidth force control, while having more torque than direct-drive actuators.
We showed that the belt reduction and differential gear system can pair well with the quasi-direct-drive actuators, as joint torque is shared between the two motors, thereby increasing the maximum possible torque at each joint of fingers. In this way, we retain most of the benefits of direct-drive actuation while increasing the possible gripping force.

Many robotic hands lack in-hand manipulation capability due to the complex motion planning and control of contact forces required with more traditional methods. Our hand achieves force control with minimal sensing (only current sensing) and simplifies motion planning by using impedance control as the relation between force and position. Perhaps this philosophy might lower the barrier for entry for performing more complicated tasks with robot hands.

While the QDD hand was successful in a lot of ways, it certainly has its limitations. Below is a list of potential design improvements for future iterations of the hand:

\begin{itemize}
  \item Reduce the size of the base structure housing the motors.
  \item Reduce the size of electronic hardware (mainly motor drivers) so they can be integrated into the structure of the hand. Ideally, we would make a custom motor driver with integrated MCU and encoder to limit the amount of wires needed and reduce the size.
  \item Design a thermal management system for the motors to increase their torque output without overheating.
  \item Add a third finger to the hand to increase capability and increase grasp stability.
\end{itemize}

Moreover, we plan on mounting the hand to the Franka Emika Panda robot arm. This will allow us to test further the ability of the hand for more complicated tasks. For example, it would be useful to know the success rate of the \textit{Smack-and-Snatch} manipulation task for many trials with different objects. That being said, with further development, we believe impedance-controlled QDD actuators show great promise in enabling more capable robot hands of the future.




\label{sec:conclution}




\bibliographystyle{IEEEtran}
\bibliography{References}

\end{document}